%% file: main.tex
\documentclass[lettersize,journal]{IEEEtran}
\usepackage{amsmath,amsfonts}
\usepackage{algorithmic}
\usepackage{algorithm}
\usepackage{array}
\usepackage{textcomp}
\usepackage{stfloats}
\usepackage{verbatim}
\usepackage{graphicx}

\usepackage[utf8]{inputenc}

\usepackage{todonotes}
\usepackage{pifont}

\usepackage{multirow}
\usepackage{booktabs}
\usepackage{caption,subcaption}
\usepackage{graphicx}
\PassOptionsToPackage{hyphens}{url}
\usepackage{xspace}
\usepackage{amssymb}
\usepackage{multirow}
\usepackage[numbers,sort&compress]{natbib}
\usepackage{wasysym}
\usepackage{hyperref}
\usepackage[hyphens]{url}
\usepackage{breakurl}
\usepackage{cleveref}
\usepackage{xr}
\usepackage{url}
\usepackage{layouts}
\usepackage{threeparttable}
\usepackage[normalem]{ulem}

\newcommand{\ie}{\textit{i.e.}\xspace}
\newcommand{\eg}{\textit{e.g.}\xspace}
\newcommand{\etc}{\textit{etc}\xspace}

\newcommand{\tabincell}[2]{\begin{tabular}{@{}#1@{}}#2\end{tabular}}
\newtheorem{definition}{Definition}

\def\major{0}

\newcommand{\highlight}[1]{%
\if\major1%
{\color{blue}#1}%
\else
#1
\fi
}

\def\minor{0}

\newcommand{\hlminor}[1]{%
\if\minor1%
{\color{blue}#1}%
\else
#1
\fi
}

\begin{document}

\title{A Survey for Federated Learning Evaluations: \\Goals and Measures}

\author{Di~Chai$^{*}$,
		Leye~Wang$^{*}$,
		Liu~Yang,
		Junxue~Zhang,
		Kai~Chen,
		QiangYang\\
		$^*$ Equal Contribution
	\IEEEcompsocitemizethanks{\IEEEcompsocthanksitem Di~Chai, Liu~Yang, Junxue~Zhang, Kai~Chen, and Qiang~Yang are with Hong Kong University of Science and Technology, Hong Kong SAR, China. Qiang~Yang is also with Webank, Shenzhen, China. E-mail: \{dchai,lyangau,zjx,kaichen,qyang\}@cse.ust.hk.
	\IEEEcompsocthanksitem Leye~Wang is with Key Lab of High Confidence Software Technologies, Ministry of Education, China, and School of Computer Science, Peking University, Beijing 100871, China. E-mail: leyewang@pku.edu.cn.
	}
}


\markboth{Journal of \LaTeX\ Class Files,~Vol.~14, No.~8, August~2021}%
{Shell \MakeLowercase{\textit{et al.}}: A Sample Article Using IEEEtran.cls for IEEE Journals}


\maketitle

\begin{abstract}
Evaluation is a systematic approach to assessing how well a system achieves its intended purpose. Federated learning (FL) is a novel paradigm for privacy-preserving machine learning that allows multiple parties to collaboratively train models without sharing sensitive data. However, evaluating FL is challenging due to its interdisciplinary nature and diverse goals, such as utility, efficiency, and security. In this survey, we first review the major evaluation goals adopted in the existing studies and then explore the evaluation metrics used for each goal. We also introduce \textit{FedEval}, an open-source platform that provides a standardized and comprehensive evaluation framework for FL algorithms in terms of their utility, efficiency, and security. Finally, we discuss several challenges and future research directions for FL evaluation.
\end{abstract}

\sloppy


\input{sections/intro}
\input{sections/criteria}
\input{sections/method}
\input{sections/fedeval-core}

\input{sections/challenge}
\input{sections/conclusion}

\section*{Acknowledgments}
The work is supported by the Key-Area Research and Development Program of Guangdong Province (2021B0101400001), the Hong Kong RGC TRS T41-603/20R, the National Key Research and Development Program of China under Grant No.2018AAA0101100.

\bibliographystyle{IEEEtranN}
\bibliography{ref.bib}

\newpage

\begin{IEEEbiography}[{\includegraphics[width=1in,height=1.25in,clip,keepaspectratio]{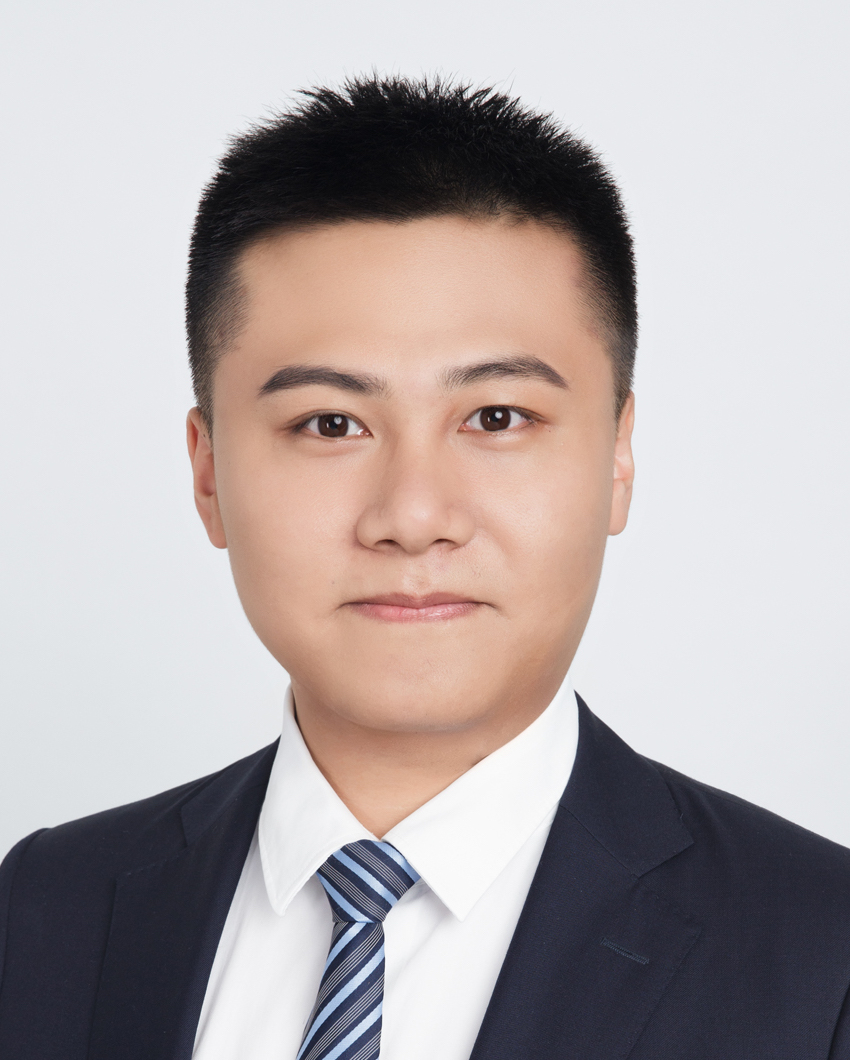}}]{Di Chai}
is a Ph.D. student in computer science and engineering at Hong Kong University of Science and Technology (HKUST). He got his master degree of science from HKUST in 2018. His research interests include federated learning and privacy-preserving machine learning.
\end{IEEEbiography}	

\vspace{-15mm}

\begin{IEEEbiography}[{\includegraphics[width=1in,height=1.25in,clip,keepaspectratio]{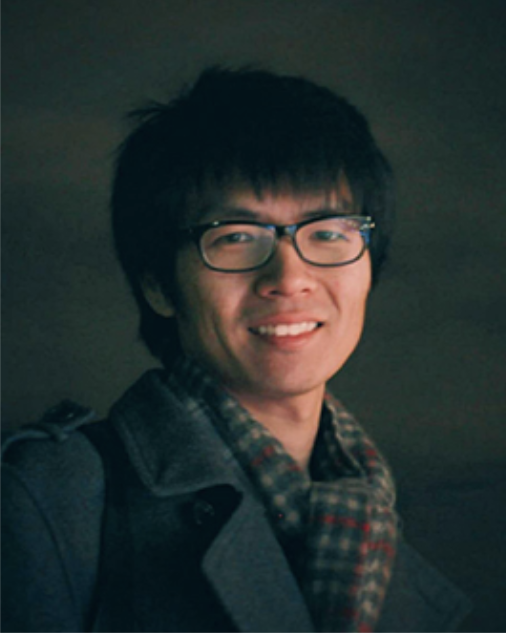}}]{Leye Wang}
	is an assistant professor at Key Lab of High Confidence Software Technologies (Peking University), MOE, and School of Computer Science, Peking University, China. He received a Ph.D. in computer science from TELECOM SudParis and University Paris 6, France, in 2016. He was a postdoc researcher with Hong Kong University of Science and Technology. His research interests include ubiquitous computing, mobile crowdsensing, and urban computing.
\end{IEEEbiography}	

\vspace{-15mm}

\begin{IEEEbiography}[{\includegraphics[width=1in,height=1.25in,clip,keepaspectratio]{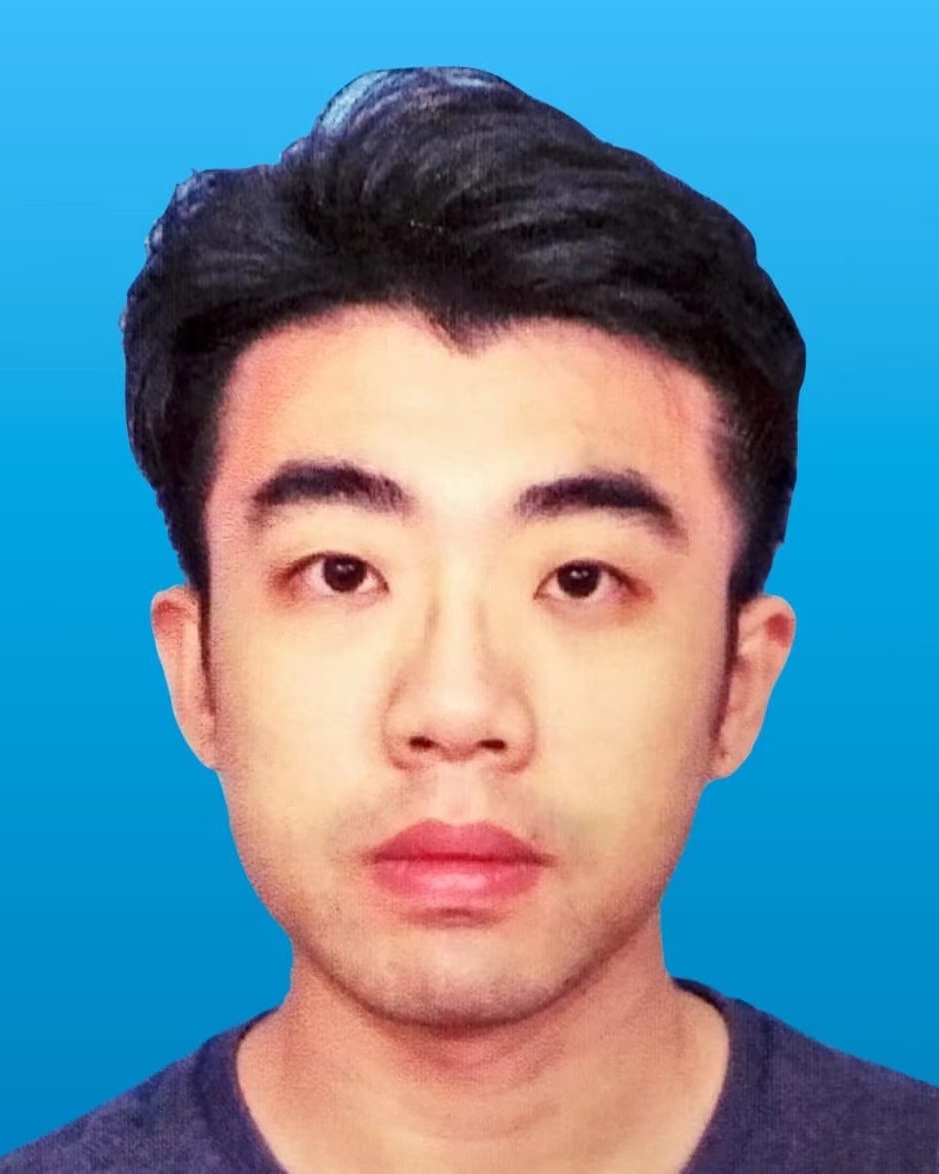}}]{Liu Yang}
	is a PhD student of computer science at iSINGLab, Hong Kong University of Science and Technology (HKUST). He is under supervision of Prof. Qiang Yang and Prof. Kai Chen. His research interests include federated learning and recommendation system. Before pursuing PhD, he received his BEng and MSc from Sun Yat-sen University and HKUST, respectively.
\end{IEEEbiography}	

\vspace{-15mm}

\begin{IEEEbiography}[{\includegraphics[width=1in,height=1.25in,clip,keepaspectratio]{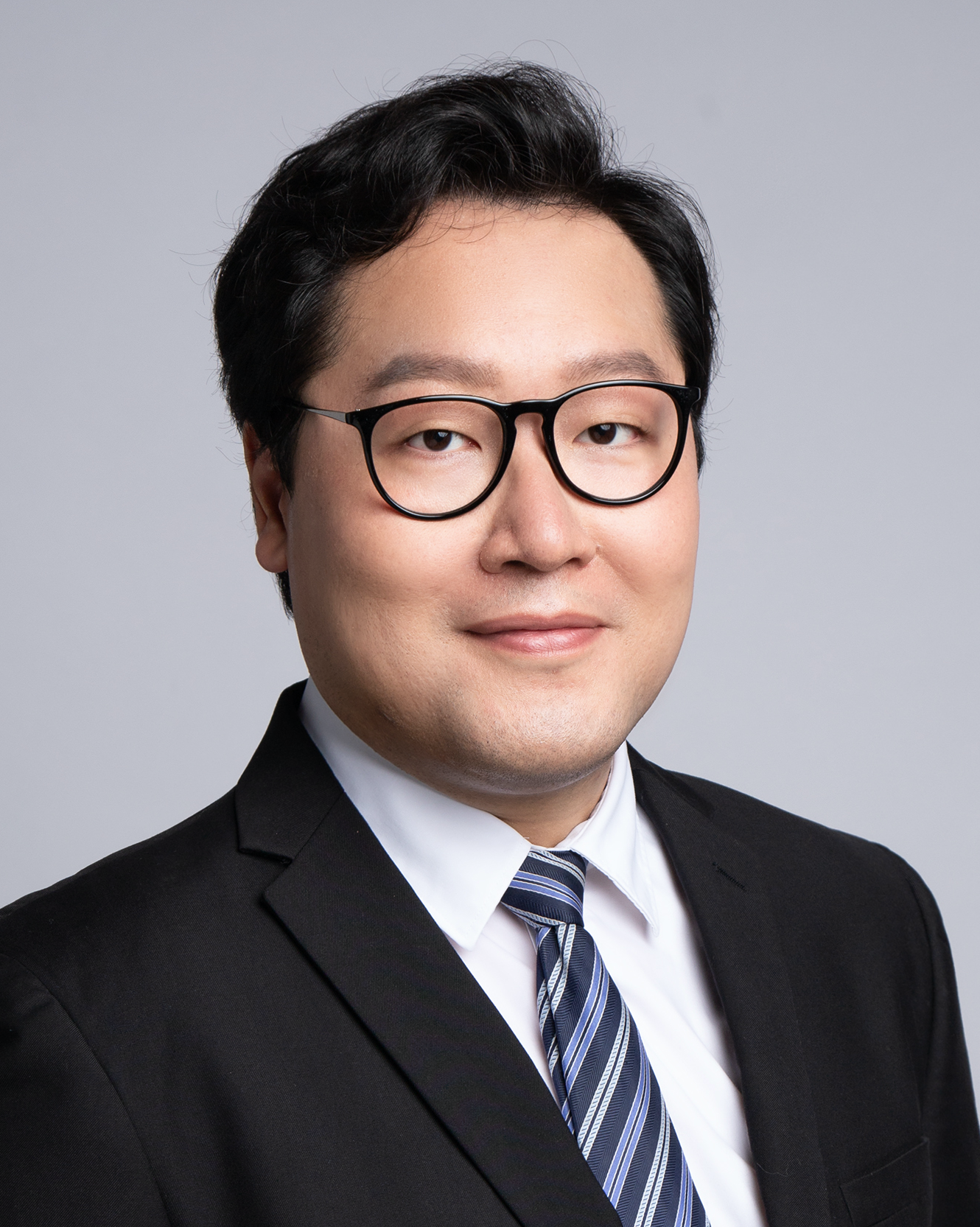}}]{Junxue Zhang}
	is currently a Research Assistant Professor at the Department of Computer Science \& Engineering, the Hong Kong University of Science and Technology (HKUST). Junxue obtained his Ph.D. at iSINGLab, HKUST, supervised by Prof. Kai CHEN. His research interests are data center networking, machine learning systems and privacy-preserving computation. His work has been published in various top venues such as SIGCOMM, NSDI and TON, including an ICNP best paper.
\end{IEEEbiography}	

\vspace{-15mm}

\begin{IEEEbiography}[{\includegraphics[width=1in,height=1.25in,clip,keepaspectratio]{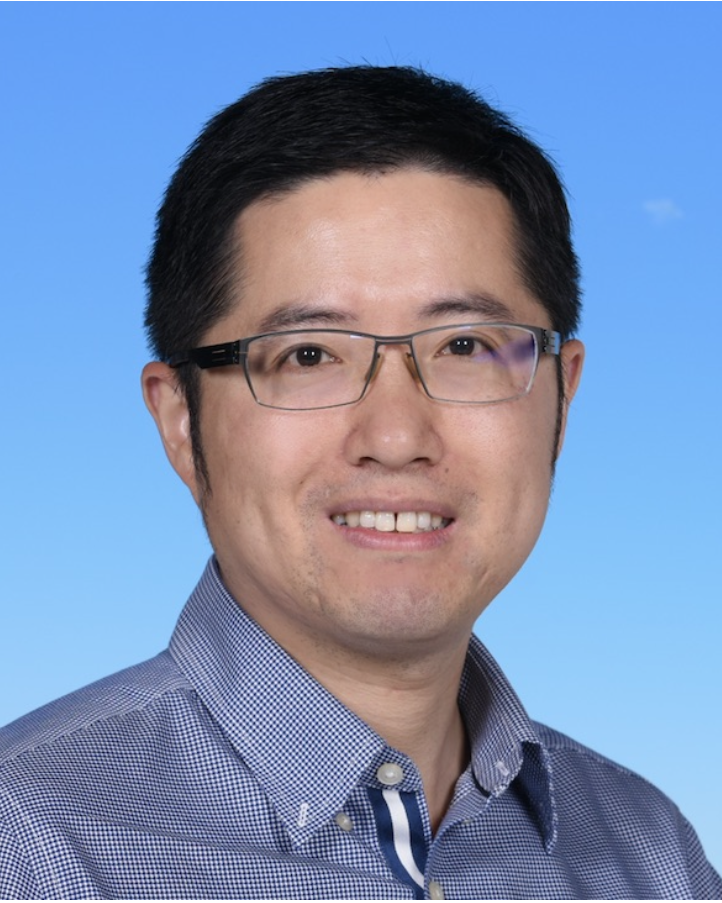}}]{Kai Chen}
	is the Professor with the Department of Computer Science and Engineering, Hong Kong University of Science and Technology, Hong Kong. He received his Ph.D. degree in Computer Science from Northwestern University, Evanston, IL, USA in 2012. His research interests include data center networking, machine learning systems and privacy-preserving computing.
\end{IEEEbiography}

\vspace{-15mm}

\begin{IEEEbiography}[{\includegraphics[width=1in,height=1.25in,clip,keepaspectratio]{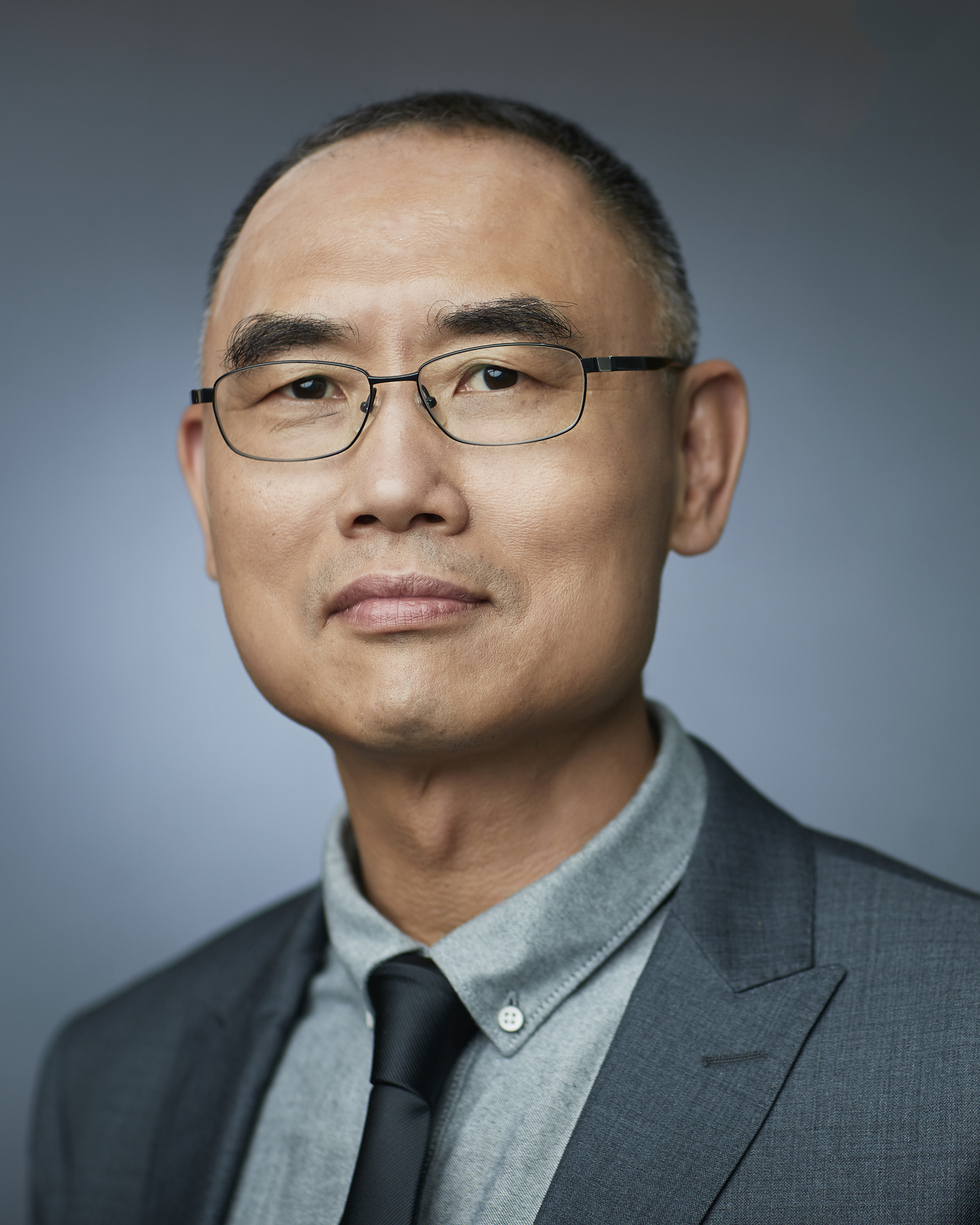}}]{Qiang Yang}
	is a Fellow of Canadian Academy of Engineering (CAE) and Royal Society of Canada (RSC), Chief Artificial Intelligence Officer of WeBank, Professor emeritus and former Chair Professor of Computer Science and Engineering Department at Hong Kong University of Science and Technology (HKUST).  He is a fellow of AAAI, ACM, IEEE, etc.  He was the Founding Editor in Chief of the ACM Transactions on Intelligent Systems and Technology (ACM TIST) and the Founding Editor in Chief of IEEE Transactions on Big Data (IEEE TBD). He was President of IJCAI from 2017 to 2019. His research interests are artificial intelligence, machine learning, data mining and planning.
\end{IEEEbiography}

\end{document}

%% file: sections/intro.tex
\section{Introduction} \label{sec:introduction}

\highlight{Federated learning (FL) is an emerging technology that aims to address data privacy concerns in real-world applications.} Data privacy has become an increasingly severe issue today as more and more real-life applications are driven by cross-domain private data. Companies that fail to protect users' privacy may face a hefty fine. For instance, the Federal Trade Commission (FTC) fined Facebook \$5 billion to force new privacy measures \cite{FTC-Fine-FB}, and Luxembourg's National Commission for Data Protection (CNPD) imposed a record-breaking fine of \$887 million on Amazon for misusing customer data for targeted advertising purposes \cite{Amazon-Fine}.
In this situation, federated learning (FL) has received many research and industry interests as a new paradigm of privacy-preserving machine learning \cite{yang2019federated}. Rather than collecting massive user data for model training, FL sets up a joint training scenario in which the clients' devices participate in model training under a joint agreement with a central authority. The client devices only upload specific model parameters to the cloud server for aggregation. Recently, FL has appeared on the Gartner `Hype Cycle for Data Science and Machine Learning' at the innovation trigger stage, indicating the importance and widespread acceptance of the FL technique \cite{GartnerCycle}.

\highlight{Evaluation plays a critical role in designing various FL algorithms and systems, owing to the need for rigorous performance assessment, providing comparative analysis between different algorithms, ensuring robustness across diverse environments, and identifying limitations for further improvement.}
Conceptually, evaluation is a systematic method to investigate how well a model, framework, or system meets its intended purposes. Essentially, two fundamental questions must be answered during the evaluation process: (1) \textit{what are the \textbf{goals} that need to be achieved?}, and (2) \textit{how can the ability to achieve these goals be \textbf{measured}?} For example, in the case of image classification, achieving \textit{high accuracy} is a primary goal; to measure accuracy, many research works have evaluated their models on the well-known public dataset, ImageNet, leading to the creation of the ImageNet leaderboard.\footnote{\url{https://paperswithcode.com/sota/image-classification-on-imagenet}}
In this article, we aim to provide clarity on the two evaluation issues for FL systems, namely goals and measurements. 
By doing so, we hope to assist researchers in conducting FL system evaluations in a more comprehensive and accessible manner and contribute to the healthy development of the entire FL community.

\highlight{The evaluation of FL is challenging as it is a multi-objective and cross-domain research topic that leverages techniques from machine learning, distributed systems, cryptography, \etc. The typical FL process usually contains three steps~\cite{yang2019federated}: 1) all parties perform local updates using private data; 2) all parties send the locally updated parameters to a third-party server, which will perform an aggregation on the received updates to produce the global updated parameter; 3) all parties download the global parameter to replace the local one and continue the next round of training. Generally, studies from the machine learning domain aim to improve the model utility, studies from distributed systems aim to improve efficiency, and privacy-preserving researchers mainly focus on privacy protections. Existing studies~\cite{NFL-1,NFL-2} have shown that these targets are not independent objectives and exhibit substantial interrelation. Enhancing one target usually has negative impacts on the other targets~\cite{NFL-1,NFL-2}. For example, increasing the number of local updates before global synchronization (\ie, reducing global synchronization frequency) can improve communication efficiency but harm model accuracy. With more local updates, a model trained on heterogeneous, non-identical-and-independent distributed (non-IID) data across clients will deviate further from the global optimum, which is known as the non-IID issue~\cite{zhao2018federated}. This illustrates the trade-off between communication efficiency and model utility, and we will discuss more trade-offs between different targets in \Cref{sec:tradeoff}. Appropriate and comprehensive evaluations can guide our future research directions by fully revealing the tradeoffs between different objectives, as well as the theoretical upper and lower bounds on the performance of different methods under varying conditions (\eg, different data distributions).
}

\highlight{Appropriate evaluation is crucial to not only promote the healthy development of FL, but the evaluations themselves can enable further applications.}

\begin{itemize}
    \item \textbf{Evaluation as Quality Control.} Real-world applications prefer FL models with excellent performance. FL models with significant issues, such as private data leakage, are unsuitable for practical applications. Therefore, FL system evaluation serves as a quality control measure for FL models before they can be used in real-world scenarios.
    \item  \textbf{Evaluation for Incentive Design.} FL system evaluation can also work with incentive mechanisms during federated training. Specifically, the contribution of each data provider needs to be quantitatively evaluated, and then the payoff of the federation can be allocated fairly according to these evaluations \cite{cong2020game,yu2020sustainable}. 
    \item \textbf{Evaluation as Online Verification.} Existing FL studies often make assumptions, particularly for security-related assumptions such as semi-honest behavior. However, these assumptions may not always hold in practice. FL system evaluation can serve as an online verification tool to ensure that FL participants adhere strictly to the pre-defined protocol.
\end{itemize}

\highlight{In contrast, the inappropriate evaluation will produce biased assessments, and the undiscovered limitations in FL algorithms or systems will damage real-world applications. For example, undiscovered privacy vulnerabilities will not only leak data providers' privacy but also decrease people's trust and willingness to further contribute data in the federated systems; FL algorithms untested under different data distributions may achieve poor model quality in applications as the data distribution in real-world applications can be highly heterogeneous~\cite{zhao2018federated,li2021fedbn,yang2021achieving,sattler2019robust,jeong2018communication,li2019convergence,DBLP:conf/iclr/YangFL21,DBLP:conf/iclr/LiJZKD21}; FL systems not evaluated on real-world network conditions may fail to achieve expected efficiency in applications due to the limited bandwidth and high latency in real-world applications.}

In this survey, we first summarize the evaluation goals for FL. We then introduce various well-studied metrics and procedures for measuring these evaluation goals. Furthermore, we will present an open-source platform for FL evaluation called \textit{FedEval}.\footnote{\url{https://github.com/Di-Chai/FedEval}} This platform can aid researchers in implementing a standardized and comprehensive FL evaluation procedure with ease. Finally, we will discuss the challenges and future directions for FL system evaluations. 

\emph{Necessity of our evaluation survey} The fast development of FL has motivated many survey studies to summarize the advances and challenges of FL. Specifically, existing FL survey studies \cite{yang2019federated,zhang2021survey,aledhari2020federated,abdulrahman2020survey,rahman2021challenges} introduced the concepts and applications of FL, \cite{zhu2021federated} emphasized the non-IID studies, \cite{mothukuri2021survey,li2021survey,yin2021comprehensive} focused on the security and privacy in FL, \cite{zhan2021survey} focused on the incentive design, \cite{nguyen2021federated,lim2020federated,khan2021federated,imteaj2021survey} emphasized the internet of things (IoT) scenario, \cite{nguyen2022federated,antunes2022federated,pfitzner2021federated} summarized the medical and health case applications of FL, \cite{jiang2020federated} and \cite{fu2022federated} introduced the application of smart city and graph learning, respectively. Existing FL surveys focus on elaborating the new techniques and applications of FL, and the survey study on the evaluation of FL has been lacking. However, the evaluation of FL is a complicated problem since FL is a cross-domain topic that consists of machine learning, distributed systems, and privacy-preserving techniques, making the evaluation of FL contains many targets, \eg, utility, robustness, privacy preservation, \etc. An unreasonable evaluation process will cause an unjustified assessment of FL methods and may bring severe issues in real-world applications, \eg, one not well-evaluated FL algorithm in the health care application can cause medical accidents. Thus, the survey study on the evaluation of FL to comprehensively analyze the evaluation targets and uncover the challenges in FL evaluation is urgently required to promote the healthy development of FL.

%% file: sections/criteria.tex
\section{Federated Learning Evaluation Goals} \label{sec:goals}

\begin{figure*}[ht]
	\centering
	\includegraphics[width=.7\linewidth]{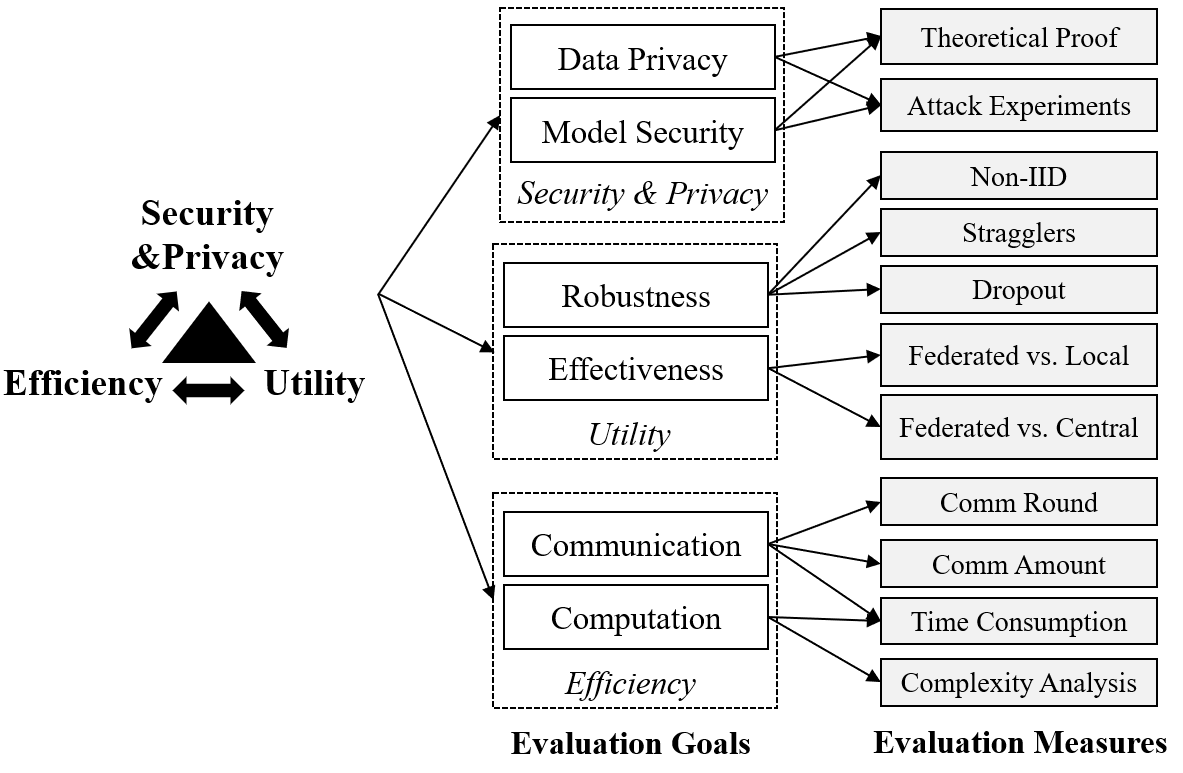}
	\caption{An overview of FL evaluation goals and measures. \highlight{Briefly, we categorize the evaluation goals of FL into three types: security \& privacy, utility, and efficiency (\Cref{sec:goals}). Then, we summarize how to measure these goals in detail (\Cref{sec:measure}).}}
	\label{fig:evaluation}
\end{figure*}

In this section, we summarize all the goals that need to be considered in the evaluation of FL (\Cref{fig:evaluation}).
In general, there are two main types of FL processes: horizontal federated learning (HFL) and vertical federated learning (VFL). HFL assumes that parties have the same feature space but different sample spaces; generally, HFL is applied in edge computing scenarios, \eg, different edge users collaboratively train the next-word-prediction model \cite{mcmahan2017communication}. 
VFL assumes that parties have the same sample space but different feature spaces; VFL is typically a to-business paradigm of FL, which happens between organizations, \eg, banks need data from online shopping companies to decide whether to approve one user's credit card application.
The evaluation goals and measures presented in this survey do not restrict the type of FL and work with both HFL and VFL.

\subsection{Goal 1: Utility}

FL generally learns a model based on data from multiple parties without directly collecting data together to meet data protection requirements in many laws and regulations. 
Hence, the primary goal is to obtain a federated model with almost the same predictive power as the model directly trained from all parties' data to ensure the high \textit{utility} of the FL model.
We discuss utility from two aspects: \textit{effectiveness} and \textit{robustness}. 

\vspace{+.5em}
\noindent \textbf{Goal 1.1: Effectiveness}. FL aims to train a global model collaboratively using data distributed across participants. Ideally, the FL training should be able to achieve the same prediction accuracy as centralized training (\ie, collecting all the data in one place).
For the FL system that can approximate a centralized model's predictive power, we then call this FL system with \textit{high effectiveness}. 

\vspace{+.5em}
\noindent \textbf{Goal 1.2: Robustness}. In practice, FL systems cannot always run in an ideal experimental environment, and various incidents may occasionally happen. 
Hence, a comprehensive evaluation of the FL system should pre-define such scenarios as much as possible to reflect the system's \textit{robustness} in practice. 
In particular, many participants indicate a significant disparity in devices. Data distributions, communication networks (3G, 4G, WiFi), computing resources (CPU, GPU), \etc, may vary among parties. These diversities and uncertainties could cause issues that significantly affect the FL system \cite{DBLP:conf/nips/SmithCST17}.

\subsection{Goal 2: Efficiency}

Unlike conventional distributed machine learning, which is carried out on different machines in one data center \cite{DBLP:journals/corr/ChenMBJ16}, FL is performed on cross-data-center machines or edge devices, which have lower networking or computing resources \cite{DBLP:conf/nips/SmithCST17}. 
Consequently, a deep neural network that could be trained in minutes using centralized machines may take hours to finish the training in FL \cite{273723}. Thus, efficiency is essential in FL and needs to be carefully evaluated. 
Based on existing works, we categorize the efficiency evaluation into two aspects: \textbf{communication} efficiency and \textbf{computation} efficiency.

\vspace{+.5em}
\noindent \textbf{Goal 2.1: Communication Efficiency}. In HFL, the edge devices have limited networking resources, \eg, low bandwidth and high latency, making the communication between the server and devices expensive \cite{DBLP:conf/nips/SmithCST17}. In VFL, the federation usually consists of machines from different data centers (\ie, from different companies). The cross-data-center communication is slow and has high latency \cite{DBLP:conf/sigmod/FuSYJXT021}. Moreover, each party, in both HFL and VFL, may join more than one federation, and the FL training tasks from different federations will compete for resources \cite{yu2020sustainable,DBLP:journals/ai/OnnT97}, making the communication efficiency issue more severe.

\vspace{+.5em}
\noindent \textbf{Goal 2.2: Computation Efficiency}. In HFL, although the edge devices tend to have more powerful hardware, they still cannot match the ability of centralized computing servers, especially when dealing with large models \cite{DBLP:conf/mobicom/NiuWTHJLWC20}. Thus, the low computation efficiency problem cannot be dismissed in the federated learning scenario. Moreover, different parties often hold distinct computation resources, which may incur significant differences in computation speed between parties \cite{li2019smartpc}. This can further impact the whole FL method's efficiency in a complicated manner.

\subsection{Goal 3: Security \& Privacy}

Security and privacy are the foundation of FL systems. HFL algorithms, \eg, FedAvg, perform aggregation on model parameters, and the risk of private data leakage can be reduced since the users’ data never leaves their devices. However, recent works have shown that gradients can reveal input data and labels \cite{NEURIPS2019_60a6c400,zhu2021rgap}. Apart from the private data leakage threats to data holders, there are also model security threats to model users. Malicious edge parties could use data poisoning or model poisoning attacks to damage or backdoor the model. Specifically, FL is often expected to achieve the following two security and privacy goals:

\vspace{+.5em}
\noindent \textbf{Goal 3.1: Data Privacy}. FL enables different parties to jointly train machine learning models without exchanging raw data, and only intermediate results are exchanged. However, recent works have shown that the intermediate results (\eg, gradients) could be used to recover FL parties' private data \cite{NEURIPS2019_60a6c400,zhu2021rgap} when no privacy-preserving techniques are adopted (\eg, homomorphic encryption), resulting in the data privacy issue.
	
\vspace{+.5em}
\noindent \textbf{Goal 3.2: Model Security}. Federated learning happens over a bunch of distributed parties (\eg, mobile devices), and there is no root of trust in existing methods, \ie, every party could be malicious from the model users' perspective. Thus, the participants could easily attack the model using poisoning methods, resulting in the model security issue \cite{fang2020local,pmlr-v108-bagdasaryan20a}.

\subsection{Trade-off between Utility, Efficiency, and Security \& Privacy}
\label{sec:tradeoff}

It is worth noting that an FL system may not simultaneously improve all the goals, including \textit{utility}, \textit{efficiency}, and \textit{security \& privacy}. \highlight{While a new algorithm improves one goal, it remains essential to comprehensively evaluate performance on other goals as well since trade-offs exist between different goals. The comprehensive analysis helps determine whether an algorithm represents unambiguous progress over state-of-the-art solutions by improving one aspect without detriment to others or gains on one goal induce losses on others, reflecting an ambiguous contribution.}
To this end, comprehensively evaluating an FL system from all three aspects becomes extremely important to deeply understand the advantages and disadvantages of FL systems (algorithms, models). \highlight{Next, we would like to demonstrate more details about the trade-offs between the goals.}

\noindent \textbf{Utility vs. Efficiency.} Federated SGD (FedSGD) and Federated Average (FedAvg) are two mostly well-known FL methods proposed by Google \cite{mcmahan2017communication}. FedSGD inherits the settings of large-batch synchronous SGD (the state-of-the-art machine learning method used in data centers). In FedSGD, all clients synchronize the gradients before updating the local model weights. In contrast, only part of the clients participate in each round of training in FedAvg and the clients perform multiple rounds of local training before the synchronization. 

FedSGD and FedAvg reveal the trade-off of utility and efficiency in FL. On the one hand, FedAvg improves communication efficiency (\ie, fewer communication rounds) by increasing clients' local training rounds before the global synchronization. On the other hand, the increased clients' local training rounds unexpectedly drift the global model away from the global optimum under heterogeneous data distributions, making FedAvg reach worse model utility than FedSGD.

Apart from FedSGD and FedAvg, there are also other FL studies that encounter the trade-off between utility and efficiency. For example, some studies utilize gradient compression techniques to improve communication efficiency \cite{sattler2019robust}; however, the model utility may decrease since only partial model parameters are synchronized.

\noindent \textbf{Efficiency vs. Security \& Privacy.} While many privacy-preserving techniques are adopted in FL to enhance privacy and security protection, there is no free lunch. 
Privacy protection generally downgrades the efficiency of the system. 

\begin{itemize}
    \item Homomorphic Encryption (HE): HE is a special encryption algorithm that enables us to perform computations directly on encrypted numbers without decryption. HE is widely applied in FL to protect the intermediate results, \eg, the gradients \cite{kim2016efficient,POSEIDON}. The encrypted numbers (\ie, ciphertext) bring the efficiency overhead in two aspects. First, the ciphertext consumes larger storage space than plaintext, which brings communication overhead. Second, the computation on ciphertext is more complicated than plaintext, which brings computation overhead.
    
    \item Secret Sharing (SS) \cite{secureml}: SS is a secure multi-party computation framework, in which different participants secretly share their data among all participants. Each participant only holds one data partition, which leaks no private information about the raw data. Basic operations, like addition and multiplication, are defined under the partitioned data, and then computations like polynomial functions could be carried out. SS mainly brings communication overhead, especially when doing multiplication \cite{DBLP:conf/ndss/Demmler0Z15}. More specially, SS is very sensitive to the networking latency.
    
    \item Secure Aggregation (SA): 
    SA is utilized in horizontal FL to combine the parameter updates from clients in a manner that protects the privacy of the individual client's local updates from a semi-honest server \cite{DBLP:conf/ccs/BonawitzIKMMPRS17}.
    SA operates in a way similar to the addition operation in SS but with the added benefit of enhancing the resilience of the aggregation process when some clients may disconnect.
    Similar to SS, SA also introduces communication overhead.
\end{itemize}

It is worth noting that, the above protection techniques can often be incorporated into various FL algorithms \cite{mcmahan2017communication,chai2020secure} to further enhance the protection level. Meanwhile, it would incur communication and/or computation overhead. Hence, in practice, the FL system designer should decide whether these extra protection methods are necessary according to the application scenario to balance efficiency and privacy protection. 

\noindent \textbf{Utility vs. Security \& Privacy.} In addition to efficiency, some privacy-preserving techniques may also degrade the utility of FL systems.

\begin{itemize}
    \item Differential Privacy (DP): a well-known privacy-preserving technique adopted in FL is differential privacy (DP) \cite{wei2020federated}. Clients locally add DP noise to the data or model to protect the private data. DP-based FL solutions reveal the trade-off between model utility and privacy. Adding more noise will have better privacy preservation, however, will significantly downgrade the model's utility.

    \item Partial Homomorphic Encryption (PHE): another case of the trade-off between model utility and security \& privacy in FL is adopting partial homomorphic encryption (PHE) in vertical federated logistic regression (LR) \cite{DBLP:journals/corr/abs-1711-10677}, in which PHE is utilized to protect the intermediate results. Since PHE cannot support non-linear functions (\eg, Sigmoid activation function), Taylor polynomials are used to approximate the non-linear functions, which bring nonnegligible loss of model utility.
\end{itemize}

\highlight{

\subsection{Necessity of comprehensively analyzing all the goals.}

Based on our survey, we highly recommend new FL algorithm or systems to perform a comprehensive analysis on all the goals, including security and privacy, utility, and efficiency, for two reasons: 1) comprehensive analysis is the foundation of a fair comparison, and 2) comprehensive analysis is the key to find all the limitations before applied in real-world applications. Specifically, the comparison between different FL studies on partial goals is unfair because different goals form trade-offs and superiority in partial goals does not mean superiority in all goals. For instance, many works do not analyze privacy protection, which will bring unfair efficiency comparisons. Because FL algorithms' efficiency varies greatly under different privacy-protection methods. For example, differential privacy (DP) and homomorphic encryption (HE) employ different privacy mechanisms and have very different efficiencies. However, claiming the DP-based method is much more efficient than the HE-based method as a major innovation is problematic without understanding their relative privacy guarantees. The major disadvantage of DP is that it harms model utility while HE does not. Comprehensive analysis is also essential to thoroughly assess one algorithm or system and discover all the limitations, such that the issue (\eg, privacy or efficiency problems) could be fixed before being applied in real-world applications.

The major challenge of performing comprehensive analysis is the workload required for evaluations. To address this, we propose two solutions: 1) We develop a standardized evaluation platform, FedEval, to produce comparable and comprehensive results while reducing evaluation workload, and the detailed is introduced \Cref{sec:fedeval}; 2) For incremental methods that only improve one or two goals based on an existing solution, another option is to analyze that the remaining goals have identical performance to prior studies that already reported comprehensive evaluation results. However, if the remaining goals were also not previously evaluated, assessments across all goals remain necessary.
}

%% file: sections/method.tex
\section{Federated Learning Evaluation Measures} \label{sec:measure}

In this section, we review existing evaluation measures for different goals, including utility, efficiency, and security \& privacy. 
For each goal, we introduce the commonly adopted evaluation measurements and factors considered in the literature. 

\subsection{Utility Evaluation Measures}

For utility evaluation, we care about the predictive power of the obtained machine learning model. Adequate data is usually an indispensable condition for achieving satisfactory prediction accuracy, especially when deep learning is applied. However, such a condition usually cannot be satisfied in the real world due to privacy-preserving restrictions. Each data owner can only access their local data, also known as the isolated data islands problem \cite{yang2019federated}. FL systems should be able to break such isolation and achieve performance, \textit{FL Effectiveness}, better than \textit{Local Effectiveness} (\ie, training model locally without joining any federations).
In FL, we typically learn the global model by solving the following problem~\cite{mcmahan2017communication}:
\begin{equation} \label{eq:fl_target}
	\min_w f(w) = \sum_{k=1}^N p_k \cdot F_k(w) = \mathbb{E}_k[F_k(w)]
\end{equation}
where $N$ is the number of clients, $p_k \ge 0$ and $\sum_k p_k = 1$. $F_k(w)$ is defined as the empirical loss over the local data samples, \ie, $F_k(w)=\frac{1}{n_k}\sum_{i=1}^{n_k}l_i(w)$ \cite{li2019fair}, where $n_k$ is the number of samples at the $k$-th party, and we set $p_k=n_k/n$ where $n=\sum_k n_k$ is the total number of samples.

\begin{definition}[FE - FL Effectiveness] \label{def:flacc}
	We define the FL effectiveness as $ \sum_{k=1}^N p_k \cdot Acc(h(w,x_k),y_k)$, where $w$ is the model parameter learned from Equation \ref{eq:fl_target}, $h(w,x_k)$ outputs a probability distribution over the classes or categories that can be assigned to $x_k \sim D_k$, $Acc$ function computes accuracy of $h(w,x_k)$ regarding the label $y_k$, and we set $p_k=n_k/n$.
\end{definition}

\begin{definition}[LE - Local Effectiveness] \label{def:LE}
	Using the same notation in Definition \ref{def:flacc}, we define the local effectiveness as $ \sum_{k=1}^N p_k \cdot Acc(h(w_k,x_k),y_k)$, where $w_k$ is the local model parameter learned by minimizing the local objective: $w_k=\arg\min_{w} F_k(w)$, and we set $p_k=n_k/n$.
\end{definition}

\begin{definition}[CE - Central Effectiveness] \label{def:CE}
We define the central effectiveness as $Acc(h(w,x), y)$, where $w$ is the model parameter trained by $\min_w F(w):=\mathbb{E}_{x \sim D}[f(w,x)]$, $x$ represents data that collected from all the clients, and $D$ is the global data distribution.\footnote{Centralized data collection and training is only an ideal experimental situation that represents a theoretical accuracy upper bound. In reality, we usually cannot put all the data in one place due to the restriction of privacy regulations.}
\end{definition}

\vspace{+.5em}
\noindent \textbf{Effectiveness}. We can compare $FE$ and $CE$/$LE$ to measure the improvement brought by FL. \highlight{The definition of central effectiveness (\ie, \Cref{def:CE}) follows accuracy definition from conventional machine learning, \ie, the ratio of correctly predicted samples in the whole evaluation dataset \cite{james2013introduction}. While the definitions of local effectiveness (LE) and FL effectiveness (FE) are more complicated since the data is distributed across the clients. Empirically, we can compute the effectiveness of each client and then aggregate all clients' results \cite{DBLP:conf/nips/LinHLZ22,DBLP:conf/nips/Yu0K0J22,li2019fair,DBLP:conf/icml/LubanaTKDM22,NEURIPS2020_24389bfe,DBLP:conf/icml/PillutlaMMRS022,DBLP:conf/icml/LinCX0GD022,jeong2022factorized,li2022soteriafl,alam2022fedrolex,DBLP:conf/icml/ChenHLCX0Z23,liu2022privacy,pmlr-v202-li23o,pmlr-v202-mai23b,pmlr-v202-baek23a,pmlr-v202-wu23z,pmlr-v202-zhang23aa}. One problem is how to set the aggregation weights, which intuitively have two approaches: uniform weights or weighted by the number of samples. Very few studies explain which approach they use in the evaluation, but we see both types of implementations when investigating the open-sourced code on GitHub (\eg, \cite{DBLP:conf/nips/LinHLZ22}\footnote{\url{https://github.com/desternylin/perfed}} used weights by sample and \cite{DBLP:conf/nips/Yu0K0J22}\footnote{\url{https://github.com/yaodongyu/TCT}} used uniform weights). Theoretically, these two types of weights are identical if all clients hold the same number of samples. However, the number of data samples held by each client could be very heterogeneous in real-world applications, making these two weights produce incompatible results. In this survey, we recommend using weights by the number of samples, the reasons are 1) weights by the number of samples matches the loss of FL \cite{mcmahan2017communication}, which is also weighted averaged by the number of training samples; 2) uniform weights could produce biased evaluation since the clients with very small amount of data may dominate the final accuracy; If the system is specifically optimized for clients with small amount of data, we recommend to report effectiveness for these clients separately, instead of mixing with other rich-data clients. In this survey, to produce standardized and compatible measures, we use weights by the number of samples to define the effectiveness of FL and local training, which is formulated in \Cref{def:flacc} and \Cref{def:LE}.}
\begin{itemize}
    \item \textit{FE vs. CE}. FL systems aim to obtain approximately the same accuracy as centralized machine learning systems, meaning that $FE \le CE$ in general cases. If $FE \approx CE$, then the FL system demonstrates no significant decline in accuracy compared to centralized learning, which is often the optimal case for an FL algorithm. 
    
    \item \textit{FE vs. LE}. For a practically useful FL system, $FE$ should be larger than $LE$, meaning that FL gets performance improvements compared to learning only on local data. If $FE \le LE$, the FL system fails to leverage the distributed knowledge to improve the model performance and should not be used in the application.
\end{itemize}

\vspace{+.5em}
\noindent \textbf{Robustness}. 
In practice, various factors may vary to impact the performance of FL systems. Hence, these factors need to be clearly configured to evaluate an FL system's utility. 

\begin{itemize}
    \item \textit{Non-IID Data \& Model Personalization}. FL aims at fitting a model to data generated by different participants. Each participant collects the data in a non-IID manner across the network. The amount of data held by each participant may also significantly differ. The non-IID issue poses challenges to the training of FL. The model will be more difficult to reach convergence under non-IID data distribution, which could be further categorized into two main types \cite{ma2022state}.
    \begin{itemize}
        \item \textit{Non-IID feature setting}: The $P(y|x)$ of different parties are the same while the $P(x)$ are different. For example, in the FEMNIST dataset, different clients hold the same label space containing the same set of symbols, but they have different handwriting styles on the same symbols.
        \item \textit{Non-IID label setting}: The $P(x|y)$ of different parties are the same while the $P(y)$ are different. For instance, in the MNIST dataset, the non-IID data is usually simulated by allocating different labels to different parties~\cite{mcmahan2017communication} such that $P(y)$ are different while the feature distributions under the same label are the same.
    \end{itemize}
    These two non-IID settings may impact model performance differently, so it is desirable to consider both of them for a robustness experiment on an FL system.
    Besides, non-IID data distribution may also lead to the necessity of \textit{model personalization}, \ie, each party attempts to learn a personalized model suitable to its local data distribution for better utility. We can measure the effectiveness of personalization by comparing a personalized model with a non-personalized (global) one.

    \item \textit{Stragglers}. FL stragglers are defined as participants that fall behind the others regarding submitting the computation results \cite{DBLP:conf/nips/SmithCST17}. FL stragglers could be caused by low computing power or small network bandwidth, which widely exist in practical FL system deployments. Suppose an FL system does not consider stragglers in its algorithm design (\eg, relying on a purely synchronous updating strategy). In that case, stragglers may bring significant utility loss to the FL system \cite{FedProx}. If the FL follows a synchronous updating strategy, the stragglers will bring large efficiency overhead to the system. In the evaluation, stragglers could be simulated using the random delay to a certain part of the participants. Then, evaluate how the system efficiency is affected by the stragglers.

    \item \textit{Dropout}. FL dropouts are defined as participants that fail to submit the computation results in training (\eg, out of battery) \cite{DBLP:conf/nips/SmithCST17}. Dropouts could be caused by networking drop-off or system out of service. Dropouts unexpectedly change the data distribution during the FL training, which may cause a convergence issue. 
    A typical way of evaluating dropouts is by simulating dropout clients in the system and observing the influence on model performance. 

\end{itemize}

\begin{table*}[!h]
	\vspace{-2mm}
	\footnotesize
    \centering
    \setlength{\tabcolsep}{0.4em}
    \renewcommand{\arraystretch}{1.1}
    \begin{tabular}{c l l c c c c}
    \toprule
    \multirow{2}{*}{\tabincell{c}{\textbf{Venues}}} & 
    \multirow{2}{*}{\tabincell{c}{\textbf{Papers}}} & 
    \multirow{2}{*}{\tabincell{c}{\textbf{\uline{Primitive Design Goals} and Keywords}}} &
    \multirow{2}{*}{\tabincell{c}{\textbf{Effecti-}\\\textbf{-veness}}} & 
    \multicolumn{3}{c}{\textbf{Robustness}} \\
    & & & & \tabincell{c}{\textit{Non-IID}} & \tabincell{c}{\textit{Straggler}} & \tabincell{c}{\textit{Dropout}} \\\midrule
    \multirow{2}{*}{\tabincell{c}{\textit{Top}\\\textit{System}}} 
    & Oort~\cite{273723} & \uline{Efficiency}, Participant Selection & $\bullet$ & $\bullet$ & $\bullet$ & $\bullet$ \\
    & SFSL~\cite{DBLP:conf/mobicom/NiuWTHJLWC20} & \uline{Privacy}, Large-Scale Edge Computing, Recommender System & $\bullet$ & $\bullet$ & $\bullet$ & $\bullet$ \\\midrule
    
    \multirow{14}{*}{\tabincell{c}{\textit{Top}\\\textit{Security}}} 
    & FLTrust~\cite{DBLP:conf/ndss/CaoF0G21} & \uline{Security}, Byzantine-robust FL & $\circ$ & $\bullet$ & $\circ$ & $\circ$ \\
    & SecAgg~\cite{DBLP:conf/ccs/BonawitzIKMMPRS17} & \uline{Privacy}, Secure Aggregation & $\circ$ & $\circ$ & $\bullet$ & $\bullet$ \\
    & Poseidon~\cite{POSEIDON} & \uline{Privacy}, Apply Fully HE in FL & $\circ$ & $\circ$ & $\circ$ & $\circ$ \\
    
    & PrivaCT~\cite{DBLP:conf/ccs/KolluriBS21} & \uline{Privacy}, Local Differential Privacy, Clustering & $\bullet$ & $\circ$ & $\circ$ & $\circ$ \\
    & Cerberus~\cite{DBLP:conf/ccs/NaseriHMSSC22} & \tabincell{c}{\uline{Utility, Privacy\&Security}, Apply FL in Security Events Prediction} & $\bullet$ & $\bullet$ & $\circ$ & $\circ$ \\
    & EIFFeL~\cite{DBLP:conf/ccs/0001GJM22} & \uline{Privacy\&Security}, SecAgg on Verified Updates & $\circ$ & $\circ$ & $\circ$ & $\bullet$ \\
    & \citet{DBLP:conf/ccs/PasquiniFA22} & \uline{Privacy}, Attack to SecAgg & $\circ$ & $\circ$ & $\circ$ & $\circ$ \\
    & DP-GDBT~\cite{DBLP:conf/ccs/MaddockC0MJ22} & \uline{Privacy}, Differentially Private GBDT & $\bullet$ & $\circ$ & $\circ$ & $\circ$ \\
    & \citet{DBLP:conf/sp/ShejwalkarHKR22} & \uline{Security}, Benchmark of Poisoning Attacks & $\bullet$ & $\bullet$ & $\circ$ & $\circ$ \\
    & Snarkblock~\cite{DBLP:conf/sp/RosenbergMM22} & \uline{Privacy}, Federated Anonymous Blocking & $\circ$ & $\circ$ & $\circ$ & $\circ$ \\
    & \citet{fang2020local} & \uline{Security}, Local Data Poisoning Attacks & $\bullet$ & $\bullet$ & $\circ$ & $\circ$ \\
    & \citet{DBLP:conf/uss/Fu0JCWG0L022} & \uline{Privacy}, Label Inference Attack, Vertical FL & $\circ$ & $\circ$ & $\circ$ & $\circ$ \\
    & FLDP~\cite{DBLP:conf/uss/StevensSVRCN22} & \uline{Privacy}, Efficiency, Differentially Private SecAgg & $\circ$ & $\circ$ & $\circ$ & $\bullet$ \\
    & FLAME~\cite{DBLP:conf/uss/NguyenRCYMFMMMZ22} & \uline{Security}, Defending Backdoor Attacks & $\bullet$ & $\bullet$ & $\circ$ & $\circ$ \\\midrule
    
    \multirow{18}{*}{\tabincell{c}{\textit{Top}\\\textit{Database}}}
    & Refiner~\cite{DBLP:journals/pvldb/ZhangDMYJ0S021} & \uline{Security}, Incentive-Driven FL & $\circ$ & $\circ$ & $\circ$ & $\circ$ \\
    & Frog~\cite{DBLP:journals/pvldb/LiuWFWW21} & \uline{Privacy, Utility, Efficiency}, Federated Debugging & $\circ$ & $\circ$ & $\circ$ & $\circ$ \\
    & FedGraph~\cite{DBLP:journals/pvldb/YuanMWZW21} & \uline{Efficiency}, Federated Subgraph Matching & $\bullet$ & $\bullet$ & $\circ$ & $\circ$ \\
    & PFA~\cite{DBLP:journals/pvldb/LiuLXLM21} & \uline{Utility, Efficiency}, Heterogeneous Differential Privacy & $\bullet$ & $\bullet$ & $\circ$ & $\circ$ \\
    & FML~\cite{DBLP:journals/pvldb/LiDZLZ21} & \uline{Privacy}, Federated Matrix Factorization, Recommender System & $\circ$ & $\circ$ & $\circ$ & $\bullet$ \\
    & CELU-VFL~\cite{DBLP:journals/pvldb/FuMJXC22} & \uline{Efficiency}, Vertical FL & $\circ$ & $\circ$ & $\circ$ & $\circ$ \\
    & SMM~\cite{DBLP:journals/pvldb/BaoZXYOTA22} & \uline{Privacy, Utility}, Mixing DP with MPC & $\bullet$ & $\circ$ & $\circ$ & $\circ$ \\
    & OpBoost~\cite{DBLP:journals/pvldb/LiHL0PH0Q22} & \uline{Utility, Privacy}, Optimizing DP for VFL & $\circ$ & $\circ$ & $\circ$ & $\circ$ \\
    & VF$^2$Boost~\cite{DBLP:conf/sigmod/FuSYJXT021} & \uline{Efficiency}, Efficient Vertical Federated GBDT & $\bullet$ & $\circ$ & $\circ$ & $\circ$ \\
    & BlindFL~\cite{DBLP:conf/sigmod/FuXCT022} & \uline{Privacy, Utility}, Support Various kinds of Features in VFL & $\bullet$ & $\circ$ & $\circ$ & $\circ$ \\
	
    & \citet{DBLP:journals/pacmmod/Xiang0L023} & \uline{Privacy, Security}, Differentially-private and Byzantine-robust FL & $\bullet$ & $\bullet$ & $\circ$ & $\circ$ \\
    & FEAST~\cite{DBLP:journals/pacmmod/FuWX023} & \uline{Utility, Efficiency}, Federated Feature Selection & $\bullet$ & $\circ$ & $\circ$ & $\circ$ \\
    & \citet{DBLP:journals/pvldb/LiWL23} & \uline{Privacy}, Differential Private Vertical Federated Clustering & $\bullet$ & $\circ$ & $\circ$ & $\circ$ \\
    
    & FedDSR~\cite{DBLP:journals/tkde/HuangLLHJW23} & \uline{Privacy, Utility}, Federated Deep Reinforcement Learning & $\bullet$ & $\circ$ & $\circ$ & $\circ$ \\
    & MGFNAS~\cite{DBLP:journals/tkde/PanHTLHL23} & \uline{Privacy}, Federated Neural Architecture Search & $\bullet$ & $\bullet$ & $\circ$ & $\circ$ \\
    & \citet{DBLP:journals/tkde/ZhangZXZY23} & \uline{Privacy, Security}, Incentive, Game-Theoretical FL & $\bullet$ & $\circ$ & $\circ$ & $\circ$ \\
    & DSANLS~\cite{DBLP:journals/tkde/QianTDLM22} & \uline{Privacy, Efficiency}, Federated Nonnegative Matrix Factorization & $\bullet$ & $\circ$ & $\circ$ & $\circ$ \\
    & VERTICOX~\cite{DBLP:journals/tkde/DaiJBLXO22} & \uline{Utility}, Federated Survival Analysis & $\bullet$ & $\circ$ & $\circ$ & $\circ$ \\
    
    \midrule
    
    \multirow{21}{*}{\tabincell{c}{\textit{Top}\\\textit{AI}}} 
    & q-FFL~\cite{li2019fair} & \uline{Utility}, Fair Resource Allocation in FL & $\bullet$ & $\bullet$ & $\circ$ & $\circ$ \\
    & Per-FedAvg~\cite{NEURIPS2020_24389bfe} & \uline{Utility}, Personalized FL & $\bullet$ & $\bullet$ & $\circ$ & $\circ$ \\
    & pFedMe~\cite{NEURIPS2020_f4f1f13c} & \uline{Utility}, Personalized FL & $\bullet$ & $\bullet$ & $\circ$ & $\circ$ \\
    & HeteroFL~\cite{diao2021heterofl} & \uline{Efficiency}, FL for Heterogeneous Clients & $\bullet$ & $\bullet$ & $\circ$ & $\circ$ \\
    & FedMix~\cite{yoon2021fedmix} & \uline{Utility}, Mixup for FL, Data Augmentation & $\bullet$ & $\bullet$ & $\circ$ & $\circ$ \\ 
    
    & PartialFed~\cite{DBLP:conf/nips/SunHYB21} & \uline{Utility}, Cross-domain Personalized FL & $\bullet$ & $\bullet$ & $\circ$ & $\circ$ \\ 
    & FRL~\cite{DBLP:conf/nips/ParkHKSM21} & \uline{Efficiency, Utility}, Constructing Initial Model for FL via Meta Learning & $\bullet$ & $\bullet$ & $\circ$ & $\circ$ \\ 
    & \citet{DBLP:conf/icml/PillutlaMMRS022} & \uline{Utility}, Convergence Analysis & $\bullet$ & $\bullet$ & $\circ$ & $\circ$ \\ 
    & Orchestra~\cite{DBLP:conf/icml/LubanaTKDM22} &\uline{Utility, Efficiency}, Unsupervised FL & $\bullet$ & $\bullet$ & $\circ$ & $\circ$ \\ 
    & FedPU~\cite{DBLP:conf/icml/LinCX0GD022} & \uline{Utility}, FL with Positive and Unlabeled Data & $\bullet$ & $\bullet$ & $\bullet$ & $\circ$ \\ 

    & FactorizedFL~\cite{jeong2022factorized} & \uline{Utility}, Personalized FL, Parameter Factorization & $\bullet$ & $\bullet$ & $\circ$ & $\circ$ \\
    & SoteriaFL~\cite{li2022soteriafl} & \uline{Privacy, Efficiency}, Differentially Private FL, Communication Compression & $\bullet$ & $\circ$ & $\circ$ & $\circ$ \\
    & FedRolex~\cite{alam2022fedrolex} & \uline{Utility}, Model-Heterogeneous FL & $\bullet$ & $\bullet$ & $\circ$ & $\circ$ \\
    & FedNTD~\cite{lee2022preservation} & \uline{Utility}, Forgetting Issues in FL, Continual Learning & $\bullet$ & $\bullet$ & $\circ$ & $\circ$ \\
    & MR-MTL~\cite{liu2022privacy} & \uline{Privacy, Utility}, Differentially Private Cross-silo FL & $\bullet$ & $\bullet$ & $\circ$ & $\circ$ \\
    
    & Fed-EF~\cite{pmlr-v202-li23o} & \uline{Efficiency}, Utility, Compressed FL with Error Feedback & $\bullet$ & $\bullet$ & $\circ$ & $\circ$ \\
    & VerFedGNN~\cite{pmlr-v202-mai23b} & \uline{Utility}, Vertical Federated Graph Neural Network & $\bullet$ & $\circ$ & $\circ$ & $\circ$ \\
    & FED-PUB~\cite{pmlr-v202-baek23a} & \uline{Utility}, Personalized Sub-graph FL & $\bullet$ & $\bullet$ & $\circ$ & $\circ$ \\
    & FedGMM~\cite{pmlr-v202-wu23z} & \uline{Utility}, Improving Effectiveness of FL on Unseen Data & $\bullet$ & $\bullet$ & $\circ$ & $\circ$ \\
    & GuardHFL~\cite{DBLP:conf/icml/ChenHLCX0Z23} & \uline{Privacy, Efficiency}, Heterogeneous Client Capabilities, Customized Model & $\bullet$ & $\bullet$ & $\circ$ & $\circ$ \\
    & PFL~\cite{pmlr-v202-zhang23aa} & \uline{Efficiency}, Asynchronized and Parallel FL  & $\bullet$ & $\bullet$ & $\bullet$ & $\bullet$ \\
    \bottomrule
	
 \end{tabular}
	\caption{Utility evaluations in recent representative FL papers. \highlight{To better identify the characteristics of each work, we present the papers' system names, primitive design goals, and keywords, which are summarized based on the papers' abstract and introduction. We use the authors' names as substitutes if the paper does provide a system name (\eg, \citet{DBLP:conf/ccs/PasquiniFA22}). In the table, the black and white dots indicate whether the research work considers the corresponding measurements in the evaluation or not, which is investigated from the evaluation sections of the paper.}}
	\label{tab:utility}
	\vspace{-2mm}
\end{table*}

\noindent \textbf{Existing Works on Utility Evaluation.} Table \ref{tab:utility} outlines representative FL studies and their evaluation measures for utility.
Our analysis reveals that most studies have at least one experiment focused on utility, such as comparing FL prediction accuracy with centralized, local, or other baseline FL methods' prediction accuracy.
This is particularly true for papers published in database and AI conferences, where utility is usually the primary evaluation goal. Meanwhile, regarding the robustness evaluation, most of the AI studies focused on evaluating the performance under the non-IID data and overlooked the evaluation of heterogeneous systems, \ie, when systems contain stragglers and dropouts. Specifically, only two papers \cite{DBLP:conf/icml/LinCX0GD022,pmlr-v202-zhang23aa} evaluated the heterogeneous system in the surveyed representative studies. Experiments on straggler and dropout impact primarily appear in system papers \cite{273723,DBLP:conf/mobicom/NiuWTHJLWC20}, while non-IID issues are mainly addressed by AI papers. One major reason is that the impact of non-IID data is usually modeled as a learning problem \cite{FedProx,NEURIPS2020_24389bfe,NEURIPS2020_f4f1f13c,diao2021heterofl,pmlr-v202-wu23z}, and various solutions are proposed by AI studies. However, the system heterogeneity is an essential challenge in FL since real-world FL applications usually deal with millions of clients, making it challenging to coordinate \cite{273723,DBLP:conf/mobicom/NiuWTHJLWC20}, and the heterogeneous system could decrease both efficiency and utility \cite{273723}. Thus the evaluation of heterogeneous systems is overlooked by existing studies and should be strengthened in future studies.

\subsection{Efficiency Evaluation Measures}

Since efficiency entails both communication and computation aspects, we provide an overview of their respective measures one by one.

\vspace{+.5em}
\noindent \textbf{Communication}. Communication efficiency evaluation usually involves the following two metrics:

\begin{itemize}
    \item \textit{Communication Round (CR)}: CR measures how many rounds of communication are needed to jointly train a machine learning model from scratch to converge. Many research works draw CR-to-Accuracy curves to compare communication efficiency \cite{mcmahan2017communication,FedProx,Wang2020Federated,reddi2020adaptive,DBLP:conf/icml/RothchildPUISB020}. In some cases if the model requires a long time to converge, we can also fix a certain number of communication rounds and compare the accuracy \cite{mcmahan2017communication,DBLP:conf/icml/PillutlaMMRS022}. For instance, we may fix the CR to 500, method $A$ has better communication rounds efficiency than method $B$ if $A$ shows higher accuracy than $B$ after 500 rounds of training.
    
    \item \textit{Communication Amount (CA)}: CA measures the amount of data transmitted during the FL training. Less CA could reduce the burden brought by the limited network bandwidth. A frequently used evaluation method is plotting the CA-to-Accuracy curve, which shows how much data is transmitted when reaching a certain model accuracy \cite{Wang2020Federated,sattler2019robust}.
\end{itemize}

\noindent \textbf{Computation}. Computation efficiency evaluation typically employs the following two measures:
\begin{itemize}
    \item \textit{Theoretical Complexity Analysis:} FL carries out a privacy-preserving distributed model training, which unavoidably brings computation overhead. For example, FedAvg brings computation overhead regarding server aggregation. Apart from the computation overhead brought by the distributed training, the widely adopted privacy-preserving techniques in FL, \eg, homomorphic encryption, also bring large computation overhead and need careful analysis \citep{flash,POSEIDON}. One fundamental method to evaluate computational efficiency is doing computation complexity analysis. Method $A$ is better than $B$ if $A$ has a lower order of computation complexity.

    
    \item \textit{Time Consumption:} Apart from the complexity analysis, experimental time consumption results are also frequently used to evaluate the efficiency of FL methods. Generally, we can draw a time-to-accuracy curve to compare the time consumption of different methods when reaching the same model performance \cite{273723,DBLP:conf/nips/SunHYB21}. It is worth noting that computation time is influenced by the software and hardware environments. 
    Some studies also report the time consumption by considering both communication and computation, \ie, the total time consumption of an FL system \cite{FedSVD}. 
    Thus, when reviewing an FL paper's time consumption results, it is crucial to comprehend how time consumption is calculated.
\end{itemize}

FL applications can involve numerous participants, such as Google's federated mobile keyboard prediction with millions of participants \cite{mcmahan2017communication}.
Hence, To evaluate the practical efficiency of an FL system, conducting large-scale participant experiments may be necessary.
An ideal solution would be to conduct experiments directly on a large number of devices, where each device represents a participant.
However, only a few research institutions have the capacity to maintain and conduct evaluations on a large number of devices. 
A practical alternative is simulating all participants using a few computing servers.
Specifically, virtual machine techniques, such as \textit{Docker} containers \cite{boettiger2015introduction}, are commonly used to simulate multiple FL participants on a single server.
It is also important to note that some efficiency measurements (\eg, time consumption) can be affected by the hardware and software used in developing and deploying the system.
Therefore, when conducting a comprehensive efficiency evaluation of FL systems, it is important to configure experiment parameters (\eg, network bandwidth) during simulation.

\begin{table}[!h]
    \footnotesize
    \centering
    \setlength{\tabcolsep}{0.3em}
    \renewcommand{\arraystretch}{1.1}
    \begin{tabular}{ c  l c  c c c c}
    \toprule
    \multirow{2}{*}{\tabincell{c}{\textbf{Venues}}} & \multirow{2}{*}{\tabincell{c}{\textbf{Papers}}} & \multirow{2}{*}{\tabincell{c}{\textbf{Scale}\\\textbf{(\# Party)}}} & \multicolumn{2}{c}{\textbf{Comm}} & \multicolumn{2}{c}{\textbf{Comp}} \\
    & & & \textit{Round} & \textit{Amount} & $O(*)$ & \textit{Time} \\
    \midrule
    \multirow{2}{*}{\tabincell{c}{\textit{Top}\\\textit{System}}} 
    & Oort~\cite{273723} & Millions & $\bullet$ & $\circ$ & $\circ$ & $\bullet$ \\
    & SFSL~\cite{DBLP:conf/mobicom/NiuWTHJLWC20} & Billions & $\bullet$ & $\bullet$ & $\bullet$ & $\circ$ \\\midrule
        
    \multirow{14}{*}{\tabincell{c}{\textit{Top}\\\textit{Security}}} 
    & FLTrust~\cite{DBLP:conf/ndss/CaoF0G21} & Hundreds & $\bullet$ & $\circ$ & $\circ$ & $\circ$ \\
    & SecAgg~\cite{DBLP:conf/ccs/BonawitzIKMMPRS17} & Hundreds & $\circ$ & $\bullet$ & $\bullet$ & $\bullet$ \\
    & Poseidon~\cite{POSEIDON} & $<$Hundred & $\circ$ & $\circ$ & $\bullet$ & $\bullet$ \\

    & PrivaCT~\cite{DBLP:conf/ccs/KolluriBS21} & Thousands & $\circ$ & $\circ$ & $\circ$ & $\circ$ \\
    & Cerberus~\cite{DBLP:conf/ccs/NaseriHMSSC22} & $<$Hundred & $\circ$ & $\circ$ & $\circ$ & $\circ$ \\
    & EIFFeL~\cite{DBLP:conf/ccs/0001GJM22} & Thousands & $\bullet$ & $\circ$ & $\circ$ & $\circ$ \\ 
    & \citet{DBLP:conf/ccs/PasquiniFA22} & $\backslash$ & $\bullet$ & $\circ$ & $\circ$ & $\circ$ \\ 
    & DP-GDBT~\cite{DBLP:conf/ccs/MaddockC0MJ22} & $\backslash$ & $\circ$ & $\circ$ & $\circ$ & $\circ$ \\
    & \citet{DBLP:conf/sp/ShejwalkarHKR22} & Thousands & $\bullet$ & $\circ$ & $\circ$ & $\circ$ \\
    & Snarkblock~\cite{DBLP:conf/sp/RosenbergMM22} & $\backslash$ & $\circ$ & $\circ$ & $\circ$ & $\bullet$ \\ 
    & \citet{fang2020local} & Hundreds & $\circ$ & $\circ$ & $\circ$ & $\circ$ \\
    & \citet{DBLP:conf/uss/Fu0JCWG0L022} & $\backslash$ & $\circ$ & $\circ$ & $\circ$ & $\circ$ \\
    & FLDP~\cite{DBLP:conf/uss/StevensSVRCN22} & Thousands & $\circ$ & $\circ$ & $\circ$ & $\bullet$ \\
    & FLAME~\cite{DBLP:conf/uss/NguyenRCYMFMMMZ22} & Hundred & $\bullet$ & $\circ$ & $\circ$ & $\bullet$ \\\midrule
    
    \multirow{18}{*}{\tabincell{c}{\textit{Top}\\\textit{DB}}} 
    & Refiner~\cite{DBLP:journals/pvldb/ZhangDMYJ0S021} & $\backslash$ & $\circ$ & $\circ$ & $\circ$ & $\circ$ \\
    & Frog~\cite{DBLP:journals/pvldb/LiuWFWW21} & $<$Hundred & $\circ$ & $\circ$ & $\circ$ & $\circ$ \\
    & FedGraph~\cite{DBLP:journals/pvldb/YuanMWZW21} & $\backslash$ & $\circ$ & $\circ$ & $\circ$ & $\bullet$ \\
    & PFA~\cite{DBLP:journals/pvldb/LiuLXLM21} & $<$Hundred & $\bullet$ & $\bullet$ & $\circ$ & $\circ$ \\
    & FML~\cite{DBLP:journals/pvldb/LiDZLZ21} & $<$Hundred & $\circ$ & $\circ$ & $\circ$ & $\circ$ \\
    & CELU-VFL~\cite{DBLP:journals/pvldb/FuMJXC22} & $<$Hundred & $\bullet$ & $\circ$ & $\circ$ & $\bullet$ \\
    & SMM~\cite{DBLP:journals/pvldb/BaoZXYOTA22} & $\backslash$ & $\circ$ & $\circ$ & $\circ$ & $\circ$ \\
    & OpBoost~\cite{DBLP:journals/pvldb/LiHL0PH0Q22} & $<$Hundred & $\circ$ & $\bullet$ & $\circ$ & $\bullet$ \\
    & VF$^2$Boost~\cite{DBLP:conf/sigmod/FuSYJXT021} & $<$Hundred & $\circ$ & $\circ$ & $\circ$ & $\bullet$ \\
    & BlindFL~\cite{DBLP:conf/sigmod/FuXCT022} & $<$Hundred & $\bullet$ & $\circ$ & $\circ$ & $\circ$ \\

    & \citet{DBLP:journals/pacmmod/Xiang0L023} & $<$Hundred & $\circ$ & $\circ$ & $\circ$ & $\circ$ \\
    & FEAST~\cite{DBLP:journals/pacmmod/FuWX023} & $<$Hundred & $\circ$ & $\bullet$ & $\circ$ & $\bullet$ \\
    & \citet{DBLP:journals/pvldb/LiWL23} & $<$Hundred & $\circ$ & $\bullet$ & $\circ$ & $\bullet$ \\
    
    & FedDSR~\cite{DBLP:journals/tkde/HuangLLHJW23} & Hundreds & $\circ$ & $\circ$ & $\circ$ & $\circ$ \\
    & MGFNAS~\cite{DBLP:journals/tkde/PanHTLHL23} & $<$Hundred & $\bullet$ & $\circ$ & $\circ$ & $\circ$ \\
    & \cite{DBLP:journals/tkde/ZhangZXZY23} & Hundreds & $\circ$ & $\circ$ & $\circ$ & $\circ$ \\
    & DSANLS~\cite{DBLP:journals/tkde/QianTDLM22} & Hundreds & $\bullet$ & $\circ$ & $\circ$ & $\bullet$ \\
    & VERTICOX~\cite{DBLP:journals/tkde/DaiJBLXO22} & $<$Hundred & $\bullet$ & $\circ$ & $\circ$ & $\bullet$ \\
    
    \midrule
    
    \multirow{21}{*}{\tabincell{c}{\textit{Top}\\\textit{AI}}} 
    & q-FFL~\cite{li2019fair} & Thousands & $\bullet$ & $\circ$ & $\circ$ & $\circ$ \\
    & Per-FedAvg~\cite{NEURIPS2020_24389bfe} & $<$Hundred & $\circ$ & $\circ$ & $\circ$ & $\circ$ \\
    & pFedMe~\cite{NEURIPS2020_f4f1f13c} & Hundreds & $\bullet$ & $\circ$ & $\circ$ & $\circ$ \\
    & HeteroFL~\cite{diao2021heterofl}& Thousands & $\bullet$ & $\bullet$ & $\circ$ & $\circ$ \\
    & FedMix~\cite{yoon2021fedmix} & Hundreds & $\bullet$ & $\circ$ & $\circ$ & $\bullet$ \\ 
    
    & PartialFed~\cite{DBLP:conf/nips/SunHYB21} & $<$Hundred & $\circ$ & $\circ$ & $\circ$ & $\bullet$ \\
    & FRL~\cite{DBLP:conf/nips/ParkHKSM21} & $<$Hundred & $\bullet$ & $\circ$ & $\circ$ & $\circ$ \\
    & \citet{DBLP:conf/icml/PillutlaMMRS022} & Thousands & $\circ$ & $\circ$ & $\circ$ & $\circ$ \\
    & Orchestra~\cite{DBLP:conf/icml/LubanaTKDM22} & Hundred & $\bullet$ & $\circ$ & $\circ$ & $\bullet$ \\
    & FedPU~\cite{DBLP:conf/icml/LinCX0GD022} & $<$Hundred & $\circ$ & $\circ$ & $\circ$ & $\circ$ \\ 

    & FactorizedFL~\cite{jeong2022factorized} & $<$Hundred & $\bullet$ & $\bullet$ & $\circ$ & $\circ$ \\
    & SoteriaFL~\cite{li2022soteriafl} & $<$Hundred & $\bullet$ & $\bullet$ & $\circ$ & $\circ$ \\
    & FedRolex~\cite{alam2022fedrolex} & $>$Thousands & $\circ$ & $\circ$ & $\circ$ & $\circ$ \\
    & FedNTD~\cite{lee2022preservation} & Hundreds & $\bullet$ & $\circ$ & $\circ$ & $\circ$ \\
    & MR-MTL~\cite{liu2022privacy} & Hundreds & $\circ$ & $\circ$ & $\circ$ & $\circ$ \\

    & Fed-EF~\cite{pmlr-v202-li23o} & Hundreds & $\bullet$ & $\bullet$ & $\circ$ & $\circ$ \\
    & VerFedGNN~\cite{pmlr-v202-mai23b} & Thousands & $\circ$ & $\bullet$ & $\circ$ & $\circ$ \\
    & FED-PUB~\cite{pmlr-v202-baek23a} & $<$Hundred & $\bullet$ & $\circ$ & $\circ$ & $\circ$ \\
    & FedGMM~\cite{pmlr-v202-wu23z} & Hundreds & $\circ$ & $\circ$ & $\circ$ & $\circ$ \\
    & GuardHFL~\cite{DBLP:conf/icml/ChenHLCX0Z23} & $<$Hundred & $\bullet$ & $\bullet$ & $\circ$ & $\bullet$ \\
    & PFL~\cite{pmlr-v202-zhang23aa} & $\backslash$ & $\bullet$ & $\circ$ & $\circ$ & $\bullet$ \\
    
    \bottomrule
    
\end{tabular}
\caption{Efficiency evaluations in existing works. $O(*)$ is the computation complexity analysis. \highlight{Black dots indicate that a given study incorporated the corresponding measure in its evaluation, while white dots denote that the paper did not include that measure. Meanwhile, we also summarize the scale of efficiency evaluation in different studies, represented by the number of clients.}}
\vspace{-2mm}
\label{tab:efficiency metrics}
\end{table}

\noindent \textbf{Existing Works on Efficiency Evaluation.} Table~\ref{tab:efficiency metrics} lists the efficiency evaluation considerations in representative studies. Most of the studies report efficiency evaluation regarding communication or computation since efficiency is an essential metric that highly affects the practicality of FL methods. It is worth noting that about 75\% of the surveyed representative FL studies do not evaluate efficiency regarding both communication and computation, which could lead to biased conclusions regarding the efficiency of FL systems. For example, communication rounds are commonly used as an efficiency metric in literature, but they may not always reflect the overall efficiency of the FL method. In particular, increasing local training rounds for every update in FedAvg \cite{mcmahan2017communication} can reduce communication rounds but may not decrease overall time consumption, as it requires more local computation time for each party \cite{chai2020fedeval}. Another example that demonstrates the necessity of considering communication and computation simultaneously in the efficiency evaluation is when comparing the efficiency of two different privacy protection techniques: SS \cite{SecureAggregation,DBLP:conf/ccs/0001GJM22,DBLP:conf/sigmod/FuXCT022,DBLP:conf/icml/ChenHLCX0Z23} and HE \cite{POSEIDON,DBLP:conf/sigmod/FuXCT022}. Intuitively, HE has higher computation complexity than SS but is more communication efficient than SS \cite{chen2021homomorphic}. Biased efficiency comparison may happen if we compare HE and SS towards only one aspect of computation and communication. Regarding the number of clients used in the evaluation, we found that $\sim$20\% of studies used thousands of clients, $\sim$20\% used hundreds of clients, and $\sim$60\% used fewer than one hundred clients.

\subsection{Security \& Privacy Evaluation Measures} 
\label{sec:security and privacy evaluation}

The evaluation of FL methods regarding security and privacy could be generally conducted from both theoretical and empirical aspects: 

\begin{itemize}
	\item Theoretical: Are there privacy proofs analyzing the security and privacy of proposed methods? 
	\item Empirical: Are there experiment results showing that the proposed methods can protect participants against existing attack methods?	
\end{itemize}

While theoretical analysis is a mathematically rigorous way of validating security and privacy protection, it is still rare in existing FL papers.\footnote{We investigated 60+ FL papers published on NeurIPS, ICML, ICLR, KDD, CCS, NDSS, OSDI, \etc. in the last five years, and found that less than 10\% provided rigorous proofs.}
In addition, security and privacy measures are typically evaluated in an adversarial manner, assuming certain types of attacks.
Common threats considered in existing literature include:

\noindent \textbf{[\textit{Data Privacy}] Data Reconstruction Attacks.} In FL, exchanging intermediate results is necessary for jointly training a machine learning model while keeping private data locally.	
Some pioneering FL studies leave these intermediate results unprotected, such as uploading local updates without protection in FedAvg \cite{mcmahan2017communication}. 
Follow-up studies have shown that raw private data could be recovered from these exchanged intermediate results, including gradients and model parameters  \cite{NEURIPS2019_60a6c400,zhu2021rgap,chai2020secure,NEURIPS2020_c4ede56b,weng2020privacy}. Moreover, malicious participants may be able to reconstruct training data using model inversion attacks with only the final FL model \cite{fredrikson2015model,hidano2018model}.
	
\noindent \textbf{[\textit{Data Privacy}] Inference Attack.}: In some cases, the intermediate training results and the final FL models are not enough to recover raw data precisely, but some sensitive attributes can still be inferred. For instance, adversaries can utilize intermediate information to train an attack model that infers whether a party/sample participates in FL model training, which is known as membership inference attack \cite{7958568,Nasr_Shokri_Houmansadr_2019}.
	
\noindent \textbf{[\textit{Data Privacy}] ID Leakage.} In VFL, directly sending sample IDs and computing the intersection could leak sensitive information about a party's customers. Hence, most VFL methods use private set intersection (PSI) for ID alignment \cite{yang2019federated}. However, PSI still leaks the sample IDs inside the intersection, revealing which users have registered accounts with other participants.

\noindent \textbf{[\textit{Model Security}] Byzantine Attacks.} Malicious parties can launch data or model poisoning attacks during the federated training process so as to downgrade the FL model's performance, which is known as \textit{Byzantine attacks} \cite{fang2020local}.
Data poisoning attacks involve injecting malicious data samples before the learning process starts, while model poisoning attacks assume that adversaries can directly manipulate the model parameters sent from FL parties to the server.
	
\noindent \textbf{[\textit{Model Security}] Backdoor Attacks.} \textit{Backdoor attacks} aim to control an FL model's prediction for an attacker-chosen subtask \cite{pmlr-v108-bagdasaryan20a}. Specifically, such attacks can cause a backdoored FL model to misclassify a data sample to an attacker-chosen label. In facial recognition applications, this could allow an attacker to generate a fake ID, posing significant security risks.
Different from Byzantine attacks, backdoor attacks aim to modify the model's behavior on a small portion of data without affecting the overall prediction accuracy significantly. Hence, backdoor attacks can be particularly challenging to detect since they often do not show up during normal FL evaluation and testing procedures.

\noindent \textbf{Threat Model.} It is worth noting that a research paper on FL usually defends against only partial attacks from the above list. It is essential to first define what are the threats (\ie, the threat model) before analyzing the security \& privacy. Typically, the following assumptions would be made for potential adversaries:

\begin{itemize}
    \item Security Definition: The security definition defines the degree of honesty of participants. Generally, two types of security definitions are used in FL studies:
    \begin{itemize}
        \item \textit{Honest But Curious (Semi-Honest)}: The honest but curious setting, also known as semi-honest, assumes that the participants strictly adhere to the pre-defined protocol but attempt to learn as much information as possible from the received messages. This setting is commonly considered in security and privacy analyses presented in FL papers.
	
        \item \textit{Malicious}: The malicious participants will not strictly follow the pre-defined protocol, and take any action to achieve their goal. To model malicious behavior during joint model training in FL, it is necessary to consider the specific threats that need to be protected against. However, defending against such parties is challenging, and only a few FL studies have considered them.
    \end{itemize}
    
    \item Collusion Party Number: The ability of an FL system's defense against attacks from a single party does not guarantee protection against collusion between multiple parties. Therefore, it is essential to consider the number of parties that could collude to conduct attacks when evaluating an FL system's privacy and security levels.
\end{itemize}

\begin{table}[!h]
    \footnotesize
    \centering
    \setlength{\tabcolsep}{0.15em}
    \renewcommand{\arraystretch}{1.08}
    \begin{tabular}{c l c c c c c c}
    \toprule
    \multirow{2}{*}{\tabincell{c}{\textbf{Venues}}} & \multirow{2}{*}{\tabincell{c}{\textbf{Papers}}} & \multicolumn{2}{c}{\tabincell{c}{\textbf{Security}\\\textbf{Definitions}}} & \multicolumn{2}{c}{\tabincell{c}{\textbf{Theoretical}\\\textbf{Proof}}} & \multicolumn{2}{c}{\tabincell{c}{\textbf{Empirical}\\\textbf{Experiments}}} \\
     & & \tabincell{c}{\textit{Semi}\\\textit{Honest}} & \tabincell{c}{\textit{Malic}\\\textit{ious}} & \tabincell{c}{\textit{Model}\\\textit{Security}} & \tabincell{c}{\textit{Data}\\\textit{Privacy}} & \tabincell{c}{\textit{Model}\\\textit{Security}} & \tabincell{c}{\textit{Data}\\\textit{Privacy}} \\\midrule
     
    \multirow{2}{*}{\tabincell{c}{\textit{Top}\\\textit{Sys}}} 
    & Oort~\cite{273723} & $\circ$ & $\circ$ & $\circ$ & $\circ$ & $\circ$ & $\circ$ \\
    & SFSL~\cite{DBLP:conf/mobicom/NiuWTHJLWC20} & $\bullet$ & $\circ$ & $\circ$ & $\bullet$ & $\circ$ & $\circ$ \\\midrule
    
    \multirow{14}{*}{\tabincell{c}{\textit{Top}\\\textit{Sec-}\\\textit{urity}}} 
    & FLTrust~\cite{DBLP:conf/ndss/CaoF0G21} & $\circ$ & $\bullet$ & $\bullet$ & $\circ$ & $\bullet$ & $\circ$ \\
    & SecAgg~\cite{DBLP:conf/ccs/BonawitzIKMMPRS17} & $\bullet$ & $\circ$ & $\circ$ & $\bullet$ & $\circ$ & $\circ$ \\
    & Poseidon~\cite{POSEIDON} & $\bullet$ & $\circ$ & $\circ$ & $\bullet$ & $\circ$ & $\circ$ \\
    
    & PrivaCT~\cite{DBLP:conf/ccs/KolluriBS21} & $\circ$ & $\circ$ & $\circ$ & $\bullet$ & $\circ$ & $\circ$ \\
    & Cerberus~\cite{DBLP:conf/ccs/NaseriHMSSC22} & $\circ$ & $\bullet$ & $\circ$ & $\circ$ & $\bullet$ & $\circ$ \\
    & EIFFeL~\cite{DBLP:conf/ccs/0001GJM22} & $\circ$ & $\bullet$ & $\bullet$ & $\circ$ & $\bullet$ & $\circ$ \\
    & \citet{DBLP:conf/ccs/PasquiniFA22} & $\circ$ & $\bullet$ & $\circ$ & $\circ$ & $\bullet$ & $\circ$ \\
    & DP-GDBT~\cite{DBLP:conf/ccs/MaddockC0MJ22} & $\bullet$ & $\circ$ & $\circ$ & $\circ$ & $\circ$ & $\circ$ \\
    & \citet{DBLP:conf/sp/ShejwalkarHKR22} & $\circ$ & $\bullet$ & $\circ$ & $\circ$ & $\bullet$ & $\circ$ \\
    & \tabincell{l}{Snarkblock~\cite{DBLP:conf/sp/RosenbergMM22}} & $\circ$ & $\circ$ & $\circ$ & $\circ$ & $\circ$ & $\circ$ \\
    & \citet{fang2020local} & $\circ$ & $\bullet$ & $\circ$ & $\circ$ & $\bullet$ & $\circ$ \\
    & \citet{DBLP:conf/uss/Fu0JCWG0L022} & $\circ$ & $\bullet$ & $\circ$ & $\circ$ & $\bullet$ & $\circ$ \\
    & FLDP~\cite{DBLP:conf/uss/StevensSVRCN22} & $\bullet$ & $\bullet$ & $\circ$ & $\bullet$ & $\circ$ & $\circ$ \\
    & FLAME~\cite{DBLP:conf/uss/NguyenRCYMFMMMZ22} & $\bullet$ & $\circ$ & $\bullet$ & $\circ$ & $\bullet$ & $\circ$ \\\midrule
    
    \multirow{18}{*}{\tabincell{c}{\textit{Top}\\\textit{DB}}} 
    & Refiner~\cite{DBLP:journals/pvldb/ZhangDMYJ0S021} & $\circ$ & $\bullet$ & $\circ$ & $\circ$ & $\circ$ & $\circ$ \\
    & Frog~\cite{DBLP:journals/pvldb/LiuWFWW21} & $\bullet$ & $\circ$ & $\circ$ & $\bullet$ & $\circ$ & $\circ$ \\
    & FedGraph~\cite{DBLP:journals/pvldb/YuanMWZW21} & $\circ$ & $\circ$ & $\circ$ & $\bullet$ & $\circ$ & $\circ$ \\
    & PFA~\cite{DBLP:journals/pvldb/LiuLXLM21} & $\circ$ & $\circ$ & $\circ$ & $\bullet$ & $\circ$ & $\circ$ \\
    & FML~\cite{DBLP:journals/pvldb/LiDZLZ21} & $\circ$ & $\circ$ & $\circ$ & $\bullet$ & $\circ$ & $\circ$ \\
    & CELU-VFL~\cite{DBLP:journals/pvldb/FuMJXC22} & $\circ$ & $\circ$ & $\circ$ & $\circ$ & $\circ$ & $\circ$ \\
    & SMM~\cite{DBLP:journals/pvldb/BaoZXYOTA22} & $\circ$ & $\circ$ & $\circ$ & $\bullet$ & $\circ$ & $\circ$ \\
    & OpBoost~\cite{DBLP:journals/pvldb/LiHL0PH0Q22} & $\circ$ & $\circ$ & $\circ$ & $\circ$ & $\circ$ & $\circ$ \\
    & VF$^2$Boost~\cite{DBLP:conf/sigmod/FuSYJXT021} & $\circ$ & $\circ$ & $\circ$ & $\circ$ & $\circ$ & $\circ$ \\
    & BlindFL~\cite{DBLP:conf/sigmod/FuXCT022} & $\bullet$ & $\circ$ & $\circ$ & $\bullet$ & $\circ$ & $\circ$ \\

    & \citet{DBLP:journals/pacmmod/Xiang0L023} & $\circ$ & $\bullet$ & $\circ$ & $\bullet$ & $\bullet$ & $\bullet$ \\
    & FEAST~\cite{DBLP:journals/pacmmod/FuWX023} & $\circ$ & $\circ$ & $\circ$ & $\bullet$ & $\circ$ & $\circ$ \\
    & \citet{DBLP:journals/pvldb/LiWL23} & $\bullet$ & $\circ$ & $\circ$ & $\bullet$ & $\circ$ & $\circ$ \\
    
    & FedDSR~\cite{DBLP:journals/tkde/HuangLLHJW23} & $\circ$ & $\circ$ & $\circ$ & $\circ$ & $\circ$ & $\circ$ \\
    & MGFNAS~\cite{DBLP:journals/tkde/PanHTLHL23} & $\bullet$ & $\circ$ & $\circ$ & $\bullet$ & $\circ$ & $\circ$ \\
    & \citet{DBLP:journals/tkde/ZhangZXZY23} & $\circ$ & $\bullet$ & $\circ$ & $\bullet$ & $\bullet$ & $\circ$ \\
    & DSANLS~\cite{DBLP:journals/tkde/QianTDLM22} & $\bullet$ & $\circ$ & $\circ$ & $\bullet$ & $\circ$ & $\circ$ \\
    & VERTICOX~\cite{DBLP:journals/tkde/DaiJBLXO22} & $\circ$ & $\circ$ & $\circ$ & $\circ$ & $\circ$ & $\circ$ \\
    
    \midrule
    
    \multirow{21}{*}{\tabincell{c}{\textit{Top}\\\textit{AI}}} 
    & q-FFL~\cite{li2019fair} & $\circ$ & $\circ$ & $\circ$ & $\circ$ & $\circ$ & $\circ$ \\
    & Per-FedAvg~\cite{NEURIPS2020_24389bfe} & $\circ$ & $\circ$ & $\circ$ & $\circ$ & $\circ$ & $\circ$ \\
    & pFedMe~\cite{NEURIPS2020_f4f1f13c} & $\circ$ & $\circ$ & $\circ$ & $\circ$ & $\circ$ & $\circ$ \\
    & HeteroFL~\cite{diao2021heterofl} & $\circ$ & $\circ$ & $\circ$ & $\circ$ & $\circ$ & $\circ$ \\
    & FedMix~\cite{yoon2021fedmix} & $\circ$ & $\circ$ & $\circ$ & $\circ$ & $\circ$ & $\circ$ \\

    & PartialFed~\cite{DBLP:conf/nips/SunHYB21} & $\circ$ & $\circ$ & $\circ$ & $\circ$ & $\circ$ & $\circ$ \\
    & FRL~\cite{DBLP:conf/nips/ParkHKSM21} & $\circ$ & $\circ$ & $\circ$ & $\circ$ & $\circ$ & $\circ$ \\
    & \citet{DBLP:conf/icml/PillutlaMMRS022} & $\circ$ & $\circ$ & $\circ$ & $\circ$ & $\circ$ & $\circ$ \\
    & Orchestra~\cite{DBLP:conf/icml/LubanaTKDM22} & $\circ$ & $\circ$ & $\circ$ & $\circ$ & $\circ$ & $\circ$ \\
    & FedPU~\cite{DBLP:conf/icml/LinCX0GD022} & $\circ$ & $\circ$ & $\circ$ & $\circ$ & $\circ$ & $\circ$ \\

    & FactorizedFL~\cite{jeong2022factorized} & $\circ$ & $\circ$ & $\circ$ & $\circ$ & $\circ$ & $\circ$ \\
    & SoteriaFL~\cite{li2022soteriafl} & $\circ$ & $\circ$ & $\circ$ & $\bullet$ & $\circ$ & $\circ$ \\
    & FedRolex~\cite{alam2022fedrolex} & $\circ$ & $\circ$ & $\circ$ & $\circ$ & $\circ$ & $\circ$ \\
    & FedNTD~\cite{lee2022preservation} & $\circ$ & $\circ$ & $\circ$ & $\circ$ & $\circ$ & $\circ$ \\
    & MR-MTL~\cite{liu2022privacy} & $\circ$ & $\circ$ & $\circ$ & $\bullet$ & $\circ$ & $\circ$ \\

    & Fed-EF~\cite{pmlr-v202-li23o} & $\circ$ & $\circ$ & $\circ$ & $\circ$ & $\circ$ & $\circ$ \\
    & VerFedGNN~\cite{pmlr-v202-mai23b} & $\bullet$ & $\circ$ & $\circ$ & $\bullet$ & $\circ$ & $\bullet$ \\
    & FED-PUB~\cite{pmlr-v202-baek23a} & $\circ$ & $\circ$ & $\circ$ & $\circ$ & $\circ$ & $\circ$ \\
    & FedGMM~\cite{pmlr-v202-wu23z} & $\circ$ & $\circ$ & $\circ$ & $\circ$ & $\circ$ & $\circ$ \\
    & GuardHFL~\cite{DBLP:conf/icml/ChenHLCX0Z23} & $\bullet$ & $\circ$ & $\circ$ & $\bullet$ & $\circ$ & $\circ$ \\
    & PFL~\cite{pmlr-v202-zhang23aa} & $\circ$ & $\circ$ & $\circ$ & $\circ$ & $\circ$ & $\circ$ \\
    
    \bottomrule
    
\end{tabular}
\caption{Privacy evaluations in existing works. \highlight{Similarly, the black and white dots represent whether the studies considered the corresponding measures in the evaluation or not, respectively. Regarding the security definition, we also summarize the threat models used in representative works (\ie, semi-honest, malicious, or not defined in the paper).}}
\vspace{-5mm}
\label{tab:privacy}
\end{table}

\noindent \textbf{Existing Works on Security \& Privacy Evaluation.} Table~\ref{tab:privacy} summarizes the security and privacy evaluation measures in representative FL papers. It is notable that papers published in security conferences prioritize security and privacy evaluations. 
In addition, database papers also give significant attention to security and privacy concerns in their method design.
Existing work mainly has two approaches to evaluate data privacy: 1) Provide theoretically proofs to show that the solutions are differentially private (\eg, \cite{DBLP:conf/ccs/MaddockC0MJ22,DBLP:conf/uss/StevensSVRCN22,DBLP:journals/pvldb/LiuLXLM21,DBLP:journals/tkde/ZhangZXZY23}) or all the intermediate results are protected by HE (\eg, \cite{POSEIDON}) and secret sharing (\eg, \cite{DBLP:conf/ccs/0001GJM22,DBLP:conf/sigmod/FuXCT022,DBLP:conf/icml/ChenHLCX0Z23}); 2) Perform empirical attack experiments to show that the solutions are secure against the state-of-the-art (SOTA) attacks (\eg, \cite{DBLP:journals/pacmmod/Xiang0L023,pmlr-v202-mai23b}). 
Regarding the model security, existing studies also explored the evaluation in two ways: 1) Provide security analysis to show solutions' ability to defend the attacks (\eg, the utility loss is bounded under the poisoning attacks \cite{DBLP:conf/ndss/CaoF0G21,DBLP:conf/ccs/0001GJM22,DBLP:conf/uss/NguyenRCYMFMMMZ22}); 2) Perform empirically poisoning attacks to show the solutions' utility loss under the attacks (\eg, \cite{fang2020local,DBLP:conf/uss/Fu0JCWG0L022,DBLP:journals/pacmmod/Xiang0L023}).
We also observe that most FL papers presented at AI conferences do not explicitly discuss security and privacy issues.
Considering that security and privacy are primary motivations for developing FL systems, we suggest that AI papers should also give more attention to these concerns.

%% file: sections/fedeval-core.tex
\section{FedEval: A Platform for FL System Evaluation}\label{sec:fedeval}
\vspace{-2mm}

\begin{figure*}[ht]
\centering
\includegraphics[width=.8\linewidth]{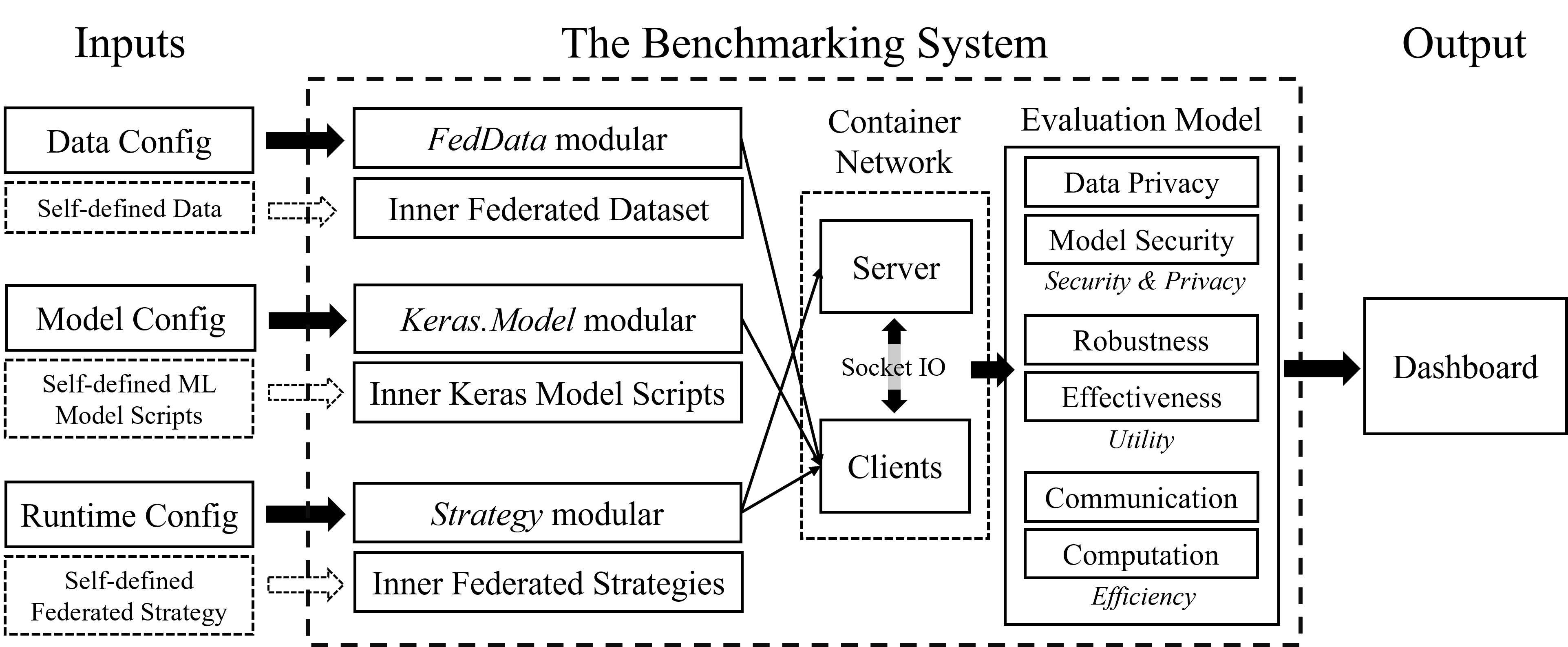}
\caption{An overview of the FedEval evaluation platform. \highlight{Users can evaluate existing algorithms using preset datasets in FedEval under different scenarios by providing the data, model, and runtime configs. Users can also evaluate new algorithms on new datasets by customizing the data, model, and strategy modules. Using the built-in evaluation goals and measures, FedEval significantly reduces the workload of the FL evaluation and produces standardized evaluation results.}}
\label{fig:bm_system_framework}
\end{figure*}

After reviewing existing FL studies, it is clear that a standard and easy-to-reproduce procedure for comprehensive evaluation of utility, efficiency, and security \& privacy is still lacking.
We have developed an open-source platform called \textit{FedEval} to standardize and simplify the evaluation of FL algorithms. An overview of our evaluation platform is presented in \Cref{fig:bm_system_framework}. To use FedEval, users only need to provide a single script that contains the necessary FL functions or callback functions, such as how the server aggregates the parameters from different clients, to evaluate a new FL algorithm or test new attack/defense methods. The platform consists of three key modules.

\begin{itemize}
    \item Data Config and the \textit{FedData} module: FedEval currently provides seven standard FL datasets, including MNIST, CIFAR10, CIFAR100, FEMNIST, CelebA, Sentiment140, and Shakespeare. Different data settings (\eg, non-IID data) can be implemented by changing the data configs. Self-defined data is also supported. We only need to inherit the \textit{FedData} class and define the \textit{load\_data} function to add a new dataset, which will share the same processing functions with the built-in datasets.
    \item Model Config and the \textit{Keras.Model} module: Currently, three machine learning models are built inside our system, including \textit{MLP}, \textit{LeNet}, and \textit{StackedLSTM}. We use TensorFlow \cite{abadi2016tensorflow} as the backend, and all the models are made via subclassing the Keras model. Thus, adding new machine learning models is very simple in FedEval.
    \item Runtime Config and the \textit{strategy} module: One of the essential components in FedEval is the \textit{strategy} module, which defines the protocol of the federated training. Briefly, the FL strategy module supports the following customization:
    \begin{itemize}
        \item \textit{Customized uploading message}, \ie, which parameters are uploaded to the server from the clients.
        \item \textit{Customized server aggregation method}, \eg, weighted average.
        \item \textit{Customized training method for clients}, \eg, the clients' model can be trained using regular gradient descent method or other solutions like knowledge distillation.
        \item \textit{Customized method for incorporating the global and local model}, \eg, one popularly used method is replacing the local model with the global one before training.
    \end{itemize}
\end{itemize}

\begin{figure*}[ht]
	\centering
	\includegraphics[width=.9\linewidth]{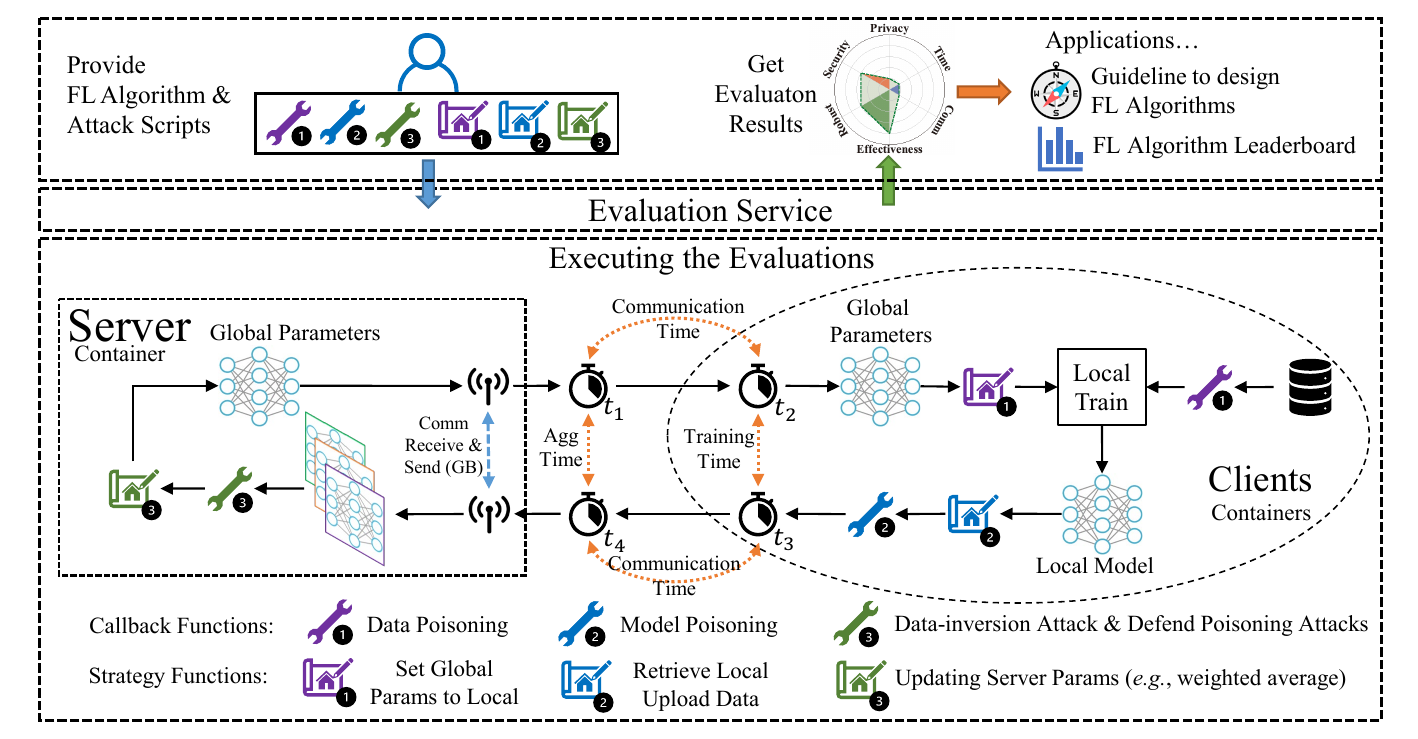}
	\caption{\hlminor{The FedEval's detailed workflow when evaluating customized algorithms. Users can provide scripts encompassing different strategy functions, enabling the assessment of various customized algorithms. For instance, these functions can customize the aggregation of parameters and the process of updating the global parameters to the local models. Additionally, users can test diverse attacking and defending techniques through different callback functions. As illustrated, clients can perform customizable data poisoning prior to local training and model poisoning before uploading updates. Conversely, the server can execute customizable data-revealing attacks and defend against poisoning attacks originating from the client side. We put the full description of the function interface of FedEval in the appendix.}}
	\label{fig:illustration_example}
\end{figure*}

\highlight{
Compared with conventional machine learning, the major challenge of obtaining standard FL evaluation metrics is how to appropriately simulate heterogeneous clients and capture metrics (\eg, communication costs) that reflect real-world conditions. We introduce the FedEval platform's approach to addressing this challenge.

\begin{itemize}
	\item Participants and Network Simulation. A widely-used method for simulating multiple participants is using multiprocessing, but we think it has the following problems: 1) it is hard to control the hardware resources (\eg, CPU and memory) used by each process; 2) it is hard to evaluate the performance under different network settings (\ie, bandwidth and latency). Our solution is putting all the participants into different docker containers, in which the hardware resources used by each participant could be fully controlled, including the CPU, GPU, memory, disk storage, \etc. The server and clients from different containers communicate through WebSocket. Container networks bridge the communication between containers. Under such an architecture design, it is easy to change the network settings (\ie, bandwidth and latency) by directly configuring the virtual network interface card (NIC).
	\item Communication Evaluation. Communication size is an essential evaluation metric for FL algorithms since the participants in FL tend to have limited network bandwidth, and a large communication size may bring significant efficiency overhead. A naive solution for evaluating the communication size, which is used in many existing FL studies, is directly measuring the size of the transmitted objects in the memory, and many utility packages (\eg, the "getsizeof()" function in Python) could be used. However, such evaluation implementation may have two issues: 1) Different packages usually have different results; 2) Not all the objects could be accurately assessed using this method. To solve these problems, we measure the communication size by directly collecting data from the virtual NIC, which automatically records the amount of data sent out and received. Compared with measuring the transmitted data size in memory, our solution is more accurate and significantly reduces the implementation complexity.
	\item \hlminor{Time Evaluation.} \hlminor{The implementation of time evaluation in FL is challenging because it may have many variations based on different purposes. For example, apart from the overall time consumption in each training round, we would also like to provide other time consumption statistics to help the users improve the FL algorithms, \eg, the computation and communication time of the clients, the aggregation time at the server, \etc. The naive implementation of these time evaluation metrics is complicated and requires significant modifications to the platform's source code. Our solution is providing a flexible time evaluation by collecting a group of timestamps, through which multiple time evaluation metrics could be calculated. Specifically, as illustrated in \Cref{fig:illustration_example}, we put four timestamps in the platform, which are the time of server sends parameters ($t_1$), clients receive parameters ($t_2$), clients send parameters ($t_3$), and server receives parameters ($t_4$). Assuming we have $k$ clients in the training, then $\{ (t_1^i, t_2^i, t_3^i, t_4^i) | 1 \le i \le k \}_n$ represents all the timestamps collected in the $i$-th round. Different combinations of these timestamps have different meanings:
	\begin{itemize}
		\item Client computation time (average): $\frac{1}{k}\sum_{i=1}^k (t_3^i - t_2^i)$.
		\item Server aggregation time in the $n$-th round:\\ $sa=min(\{t_1^i | 1 \le i \le k \}_n)-max(\{t_4^i | 1 \le i \le k \}_{n+1})$
		\item Real-world time consumption in the $n$-th round:\\ $min(\{t_1^i | 1 \le i \le k \}_n)-min(\{t_1^i | 1 \le i \le k \}_{n+1})$
		\item Federated time consumption in the $n$-th round:\\ $sa + max(\{t_4^i-t_1^i | 1 \le i \le k \}_{n})$
	\end{itemize}
	}
\end{itemize}
\hlminor{
Our platform records all the timestamps and outputs the real-world and federated time consumption. The users can compute more metrics based on these timestamps.
}
}

\highlight{With appropriate client simulation, resource control, and \hlminor{efficiency} measurements, the other metrics could be easily obtained. For example, the straggler evaluation in the utility could also be done by allocating clients with heterogeneous computing or networking resources.} The entire system is open-sourced, and the essential components, such as datasets, ML models, and FL strategies, can be easily used or self-defined.
Researchers can easily implement their new FL method ideas and evaluate them with FedEval (\eg, \textit{FedSVD} \cite{FedSVD}).


\hlminor{To demonstrate the usability of FedEval, we present its detailed workflow when evaluating customized algorithms in \Cref{fig:illustration_example}. As illustrated in the figure, the researchers can provide strategy functions to customize the behaviors of the FL algorithm, \eg, how the parameters are aggregated at the server and how to set the global updates to the local model. Meanwhile, the researchers can use customized callback functions to perform experiments of attacking and defending against the attacks. On the client side, we can use callback functions to poison the data before local training or poison the model before uploading local updates. On the server side, we can use callback functions to perform data-revealing attacks when receiving individual client updates and detect the poisoning updates before the aggregation. Due to the space limitation, we put the full description of the function interface of FedEval in the appendix.}

\begin{table}[h]
	\setlength{\tabcolsep}{0.3em}
	\renewcommand\arraystretch{0.7}
	\vspace{+5mm}
	\caption{Utility evaluation of four popular FL methods through FedEval on four datasets. All the experiments are repeated ten times, and the average values and standard error (\ie, values in parentheses) are reported. The MNIST dataset adopts the non-IID label setting, while the other datasets adopt the non-IID feature settings.}
	\label{tab:utility evaluation}
	\centering
	\begin{threeparttable}
	\begin{tabular}{c|c|c|c|c|c|c|c}
		\toprule
		Dataset & IID & Local & Central & FedSGD & FedAvg & FedProx & FedOpt  \\
		\midrule
		mnist   & N & \multirow{3}{*}{\tabincell{c}{0.11319\\(0.013)}}  & \multirow{3}{*}{\tabincell{c}{0.98614\\(0.001)}} & \tabincell{c}{0.98390\\(0.001)} & \tabincell{c}{0.97843 \\ (0.006)} & \tabincell{c}{0.97874\\(0.003)} & \tabincell{c}{0.97679\\(0.003)} \\
		\cmidrule(l){1-2}
		\cmidrule(l){5-8}
		mnist   & Y &  &  & \tabincell{c}{0.98341\\(0.002)} & \tabincell{c}{0.98651\\(0.001)} & \tabincell{c}{0.98683\\(0.001)} &  \tabincell{c}{0.98351\\(0.001)} \\
		\midrule
		femnist & N & \multirow{3}{*}{\tabincell{c}{0.48231\\(0.056)}} & \multirow{3}{*}{\tabincell{c}{0.84961\\(0.002)}} & \tabincell{c}{0.80461\\(0.015)} & \tabincell{c}{0.81234\\(0.004)} & \tabincell{c}{0.81288\\(0.005)} & \tabincell{c}{0.80783\\(0.003)} \\
		\cmidrule(l){1-2}
		\cmidrule(l){5-8}
		femnist & Y &  &  & \tabincell{c}{0.81351\\(0.012)} & \tabincell{c}{0.83476\\(0.004)} & \tabincell{c}{0.83385\\(0.002)} & \tabincell{c}{0.83187\\(0.004)} \\
		\midrule
		celebA  & N & \multirow{3}{*}{\tabincell{c}{0.70307\\(0.007)}} & \multirow{3}{*}{\tabincell{c}{0.92400\\(0.005)}} & \tabincell{c}{0.91707\\(0.005)} & \tabincell{c}{0.90170\\(0.005)} & \tabincell{c}{0.90120\\(0.007)} & \tabincell{c}{0.89913\\(0.008)} \\
		\cmidrule(l){1-2}
		\cmidrule(l){5-8}
		celebA  & Y &  &  & \tabincell{c}{0.91867\\(0.006)} & \tabincell{c}{0.90267\\(0.012)} & \tabincell{c}{0.90210\\(0.011)} & \tabincell{c}{0.89957\\(0.011)} \\
		\midrule
		sent140 & N & \multirow{3}{*}{\tabincell{c}{0.74447\\(0.006)}} & \multirow{3}{*}{\tabincell{c}{0.79263\\(0.002)}} & \tabincell{c}{0.74131\\(0.006)} & \tabincell{c}{0.75578\\(0.003)} & \tabincell{c}{0.75626\\(0.003)} & \tabincell{c}{0.75263\\(0.004)} \\
		\cmidrule(l){1-2}
		\cmidrule(l){5-8}
		sent140 & Y &  &  & \tabincell{c}{0.74024\\(0.005)} & \tabincell{c}{0.76504\\(0.004)} & \tabincell{c}{0.75839\\(0.005)} & \tabincell{c}{0.74955\\(0.007)} \\
		\midrule
		Average & N & \multirow{3}{*}{\tabincell{c}{0.51076}} & \multirow{3}{*}{\tabincell{c}{0.88809}} & \tabincell{c}{0.86172} & \tabincell{c}{0.86206} & \tabincell{c}{0.86227} & \tabincell{c}{0.85909} \\
		\cmidrule(l){1-2}
		\cmidrule(l){5-8}
		Average & Y &  &  & \tabincell{c}{0.86395} & \tabincell{c}{0.87224} & \tabincell{c}{0.87029} & \tabincell{c}{0.86612} \\
		\bottomrule
	\end{tabular}
    \end{threeparttable}
	\vspace{-4mm}
\end{table}

\begin{figure*}[ht]
    \centering
    \includegraphics[width=.99\linewidth]{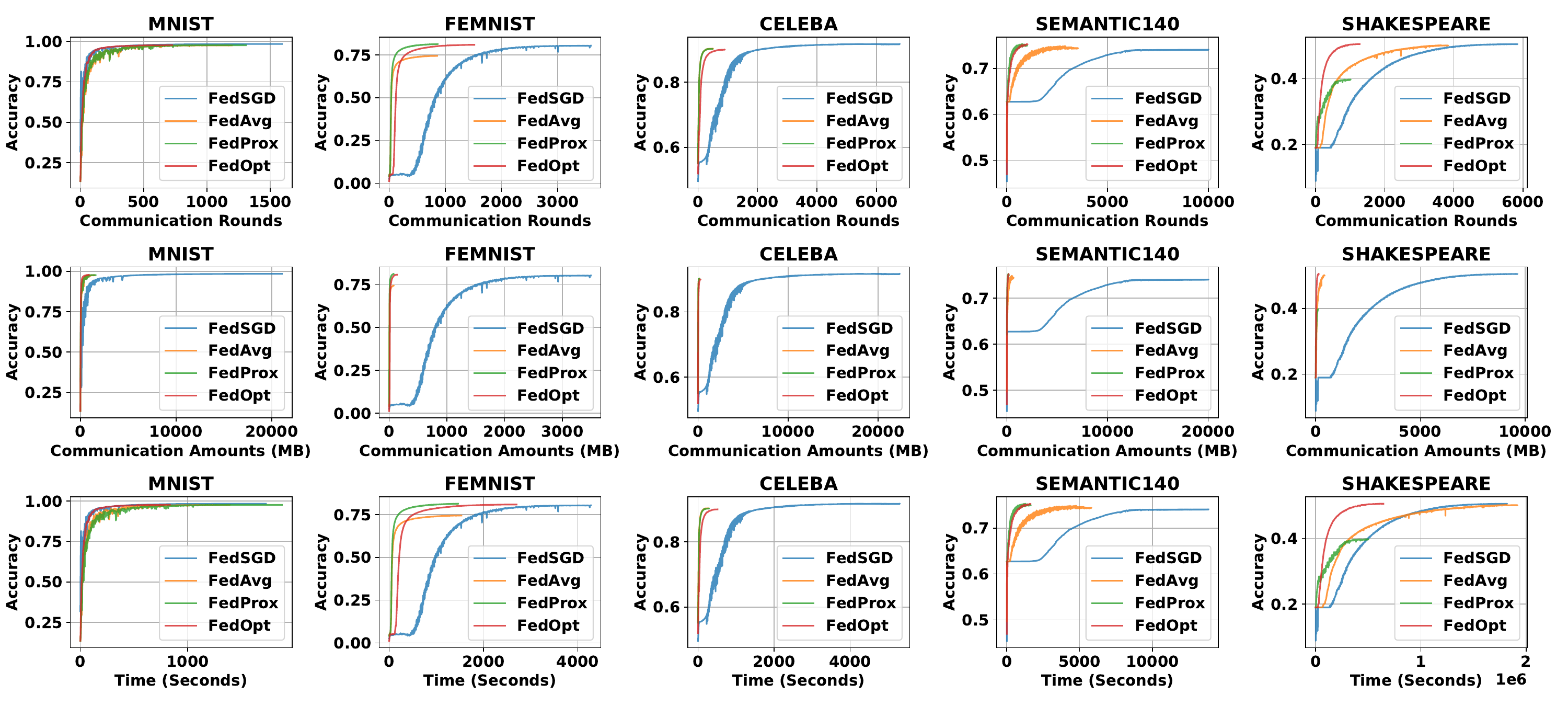}
    \caption{Efficiency evaluation of four popular FL methods through FedEval \highlight{on four datasets. The results show that FedSGD has the worst efficiency regarding both communications and computations, and FedOpt has superior efficiency on the larger dataset (\ie, Shakespeare), which match the results reported by original papers.}}
    \label{fig:fedeval_efficiency_evaluation}
\end{figure*}

\begin{figure*}[t]
	\center
	\begin{subfigure}[t]{0.22\textwidth}
		\includegraphics[width=\textwidth]{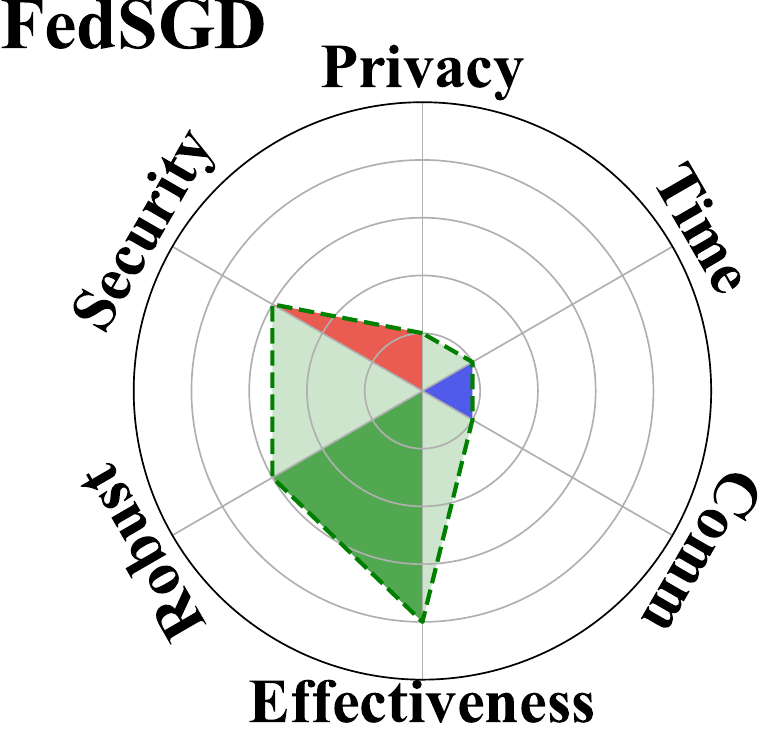} 
	\end{subfigure} \hspace{+1mm}
	\begin{subfigure}[t]{0.22\textwidth}
		\includegraphics[width=\textwidth]{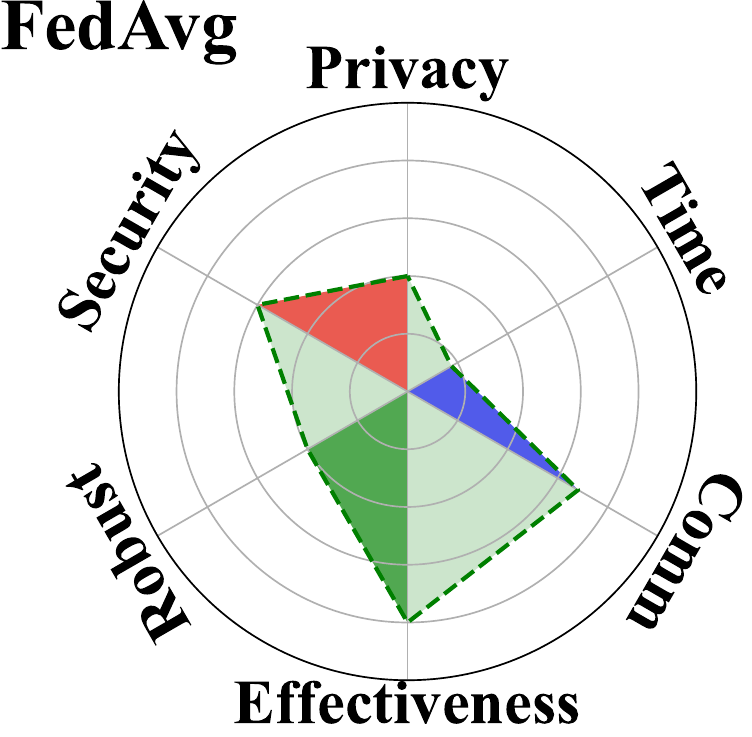}
	\end{subfigure} \hspace{+1mm}
	\begin{subfigure}[t]{0.22\textwidth}
		\includegraphics[width=\textwidth]{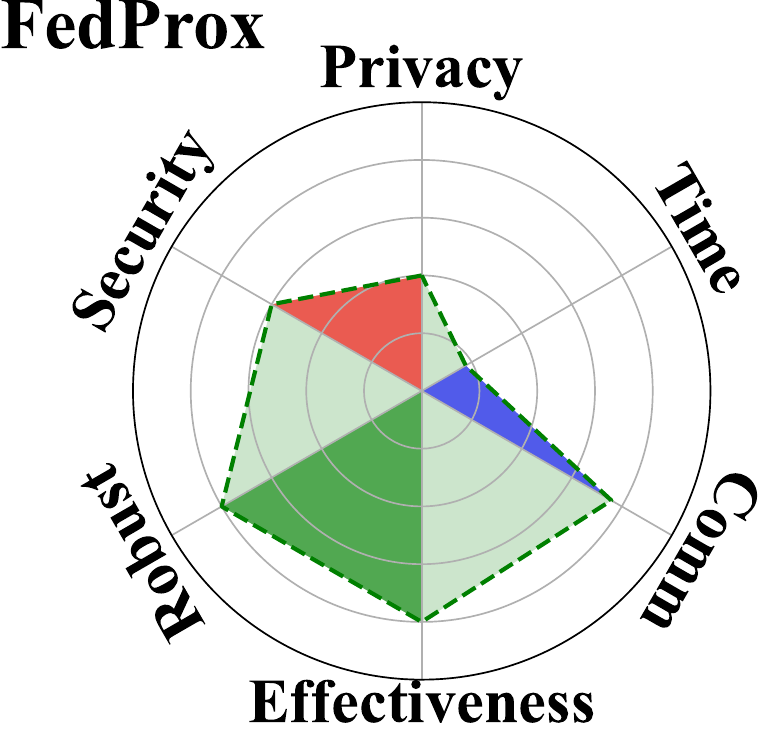}
	\end{subfigure} \hspace{+1mm}
	\begin{subfigure}[t]{0.22\textwidth}
		\includegraphics[width=\textwidth]{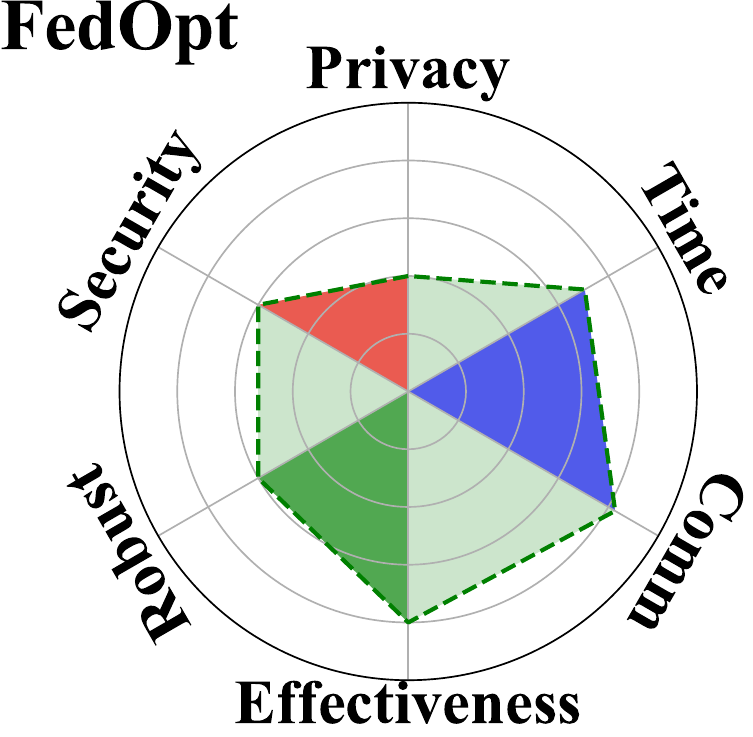}
	\end{subfigure}
	\caption{Visualizing the FedEval evaluation results through radar charts which compare four most popular FL algorithms from security and privacy, utility (\ie, robustness and effectiveness), and efficiency (\ie, communication and time consumption).}
	\label{fig:rader_charts}
	\vspace{-2mm}
\end{figure*}

\begin{figure}[h!]
	\centering
	\begin{subfigure}[t]{0.48\linewidth}
		\centering
		\includegraphics[scale=0.48]{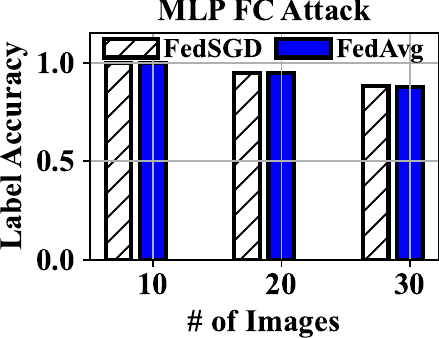}
		\label{fig:compare_fedsgd_fedavg_fc_la}
	\end{subfigure}
	\hspace{-3mm}
	\begin{subfigure}[t]{0.48\linewidth}
	\centering
		\includegraphics[scale=0.48]{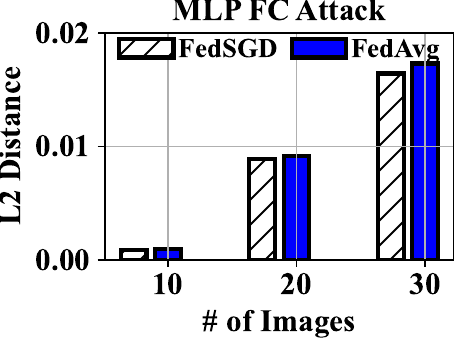}
		\label{fig:compare_fedsgd_fedavg_fc_l2d}
	\end{subfigure}
	\begin{subfigure}[t]{0.48\linewidth}
	\centering
		\includegraphics[scale=0.48]{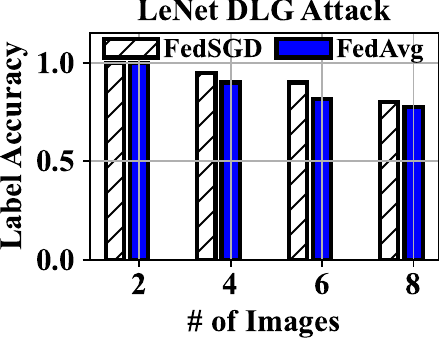}
		\label{fig:compare_fedsgd_fedavg_dlg_la}
	\end{subfigure}
	\hspace{-3mm}
	\begin{subfigure}[t]{0.48\linewidth}
	\centering
		\includegraphics[scale=0.48]{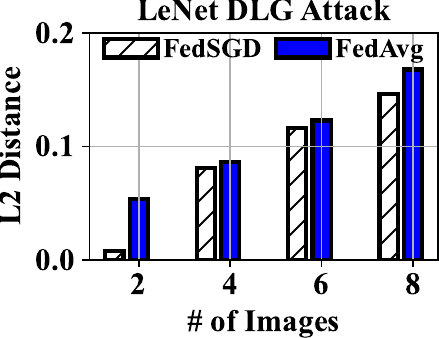}
		\label{fig:compare_fedsgd_fedavg_dlg_l2d}
	\end{subfigure}
	\vspace{-2mm}
	\caption{FedSGD vs. FedAvg under the data-reconstruction attack \cite{NEURIPS2019_60a6c400}. FedAvg has better performance than FedSGD by having lower attack label accuracy and higher L2 distance between the recovered and real samples.}
	\label{fig:comparing fedsgd and fedavg on DLG}
	\vspace{-2mm}
\end{figure}

An important characteristic of FedEval is its capability to evaluate an FL algorithm's performance from a holistic perspective including utility, efficiency, and security \& privacy. We have tested representative FL algorithms, including FedSGD~\cite{mcmahan2017communication}, FedAvg~\cite{mcmahan2017communication}, FedProx~\cite{FedProx}, FedOpt~\cite{reddi2020adaptive}, \etc. \highlight{\Cref{tab:utility evaluation} shows the utility evaluation of these four algorithms, \ie, comparing the effectiveness to local and central training and the effectiveness under non-IID data. The utility evaluation shows that all the tested FL algorithms have significantly better performance than local training and show a small decrease in accuracy compared to centralized training on most datasets. Regarding the robustness under non-IID data setting, FedProx has the best performance and yields the best average effectiveness under non-IID data, which matches the results reported from the original paper.} \Cref{fig:fedeval_efficiency_evaluation} shows the efficiency comparison of these four algorithms regarding the communication rounds, communication amounts, and time consumption. \highlight{The efficiency evaluation shows that FedSGD tends to have worse efficiency compared to the other three algorithms, and FedOpt shows superior efficiency on a relatively large dataset (\ie, Shakespeare), which also matches the results report from the original paper.} \highlight{\Cref{fig:comparing fedsgd and fedavg on DLG} shows the data reconstruction attack \cite{NEURIPS2019_60a6c400} between FedSGD and FedAvg. Theoretically, FedProx and FedOpt have the same attack results as FedAvg since clients in these protocols upload the same information (\ie, parameters after multiple rounds of local updates) to the server. \Cref{fig:comparing fedsgd and fedavg on DLG} shows that FedAvg has better performance than FedSGD. The possible reason is that the parameters uploaded in FedAvg contain multiple rounds of local training while FedSGD only has one round of training, and the accumulated local updates in the parameters make it harder to recover the raw data.}

\highlight{While the above table and figures independently present the evaluation results regarding utility, efficiency, and privacy, we also attempt to merge the evaluation results into one radar chat to provide an overview as well as highlight the strengths and weaknesses of each algorithm. The final results are presented in Figure~\ref{fig:rader_charts}. We put the detailed methods for obtaining the radar charts on an online document\footnote{\url{https://fedeval.readthedocs.io/en/latest/benchmark/benchmark.html}} due to the space limitation and ease of future updates, \ie, we will also continue evaluating more algorithms and the radar charts may also be updated accordingly. For more detail of FedEval, \eg, the interface design, please refer to our technical report \cite{chai2020fedeval} as well as the online document\footnote{\url{https://fedeval.readthedocs.io/}}.

In summary, FedEval provides a flexible framework for researchers to produce standardized evaluation results that closely mimic real-world settings using the measurements summarized in this survey. FedEval also reduces the workload required for comprehensive analysis since researchers only need to define the FL workflow (\ie, through scripts), and evaluations can be automatically completed using the built-in metrics on the platform. While being a platform that makes a significant contribution to the evaluation of FL, FedEval also has two limitations. Firstly, while the platform provides good support for utility and efficiency evaluations, the attacks for privacy and security evaluation still need to be enriched. Secondly, the automated evaluation of vertical FL algorithms is currently not supported by FedEval. We will keep updating the platform in the future to solve these two limitations, \ie, adding more attacks regarding the privacy and security evaluation and adding support for the evaluation of vertical FL.
}

%% file: sections/challenge.tex
\section{Future Directions} \label{sec:future directions}

In this section, we summarize several challenges and future research directions in FL evaluation. 

\subsection{A Comprehensive Evaluation Procedure}

While existing works focus on one or two issues in FL, their evaluation results are also restricted to the corresponding areas. For example, FedAvg \cite{mcmahan2017communication} tries to reduce the communication rounds by adding the number of clients' local updates. However, the resulting increased local running time is not evaluated; non-IID issues are not thoroughly tested. FLTrust \cite{DBLP:conf/ndss/CaoF0G21} proposed a Byzantine attack-robust FL framework by carefully verifying clients' uploaded updates; however, individual updates for verification may bring the risk of private data leakage. 
As trade-offs widely exist in FL system design (Sec.~\ref{sec:tradeoff}), only a comprehensive evaluation process can help practitioners make the optimal decision on the design of practical FL systems and applications.

\subsection{Standard Evaluation Metrics}

Although the comprehensive evaluation gives us a thorough assessment of FL frameworks, comparing different FL studies is still very difficult because the existing evaluation metrics are incompatible.
Different studies usually have different focuses in the evaluation. For example, model $A$ improves the FL communication efficiency by 10\%, and model $B$ improves the FL computation efficiency by 15\%. We cannot conclude that model $B$ is better than model $A$ and vice versa since none of these two metrics (\ie, communication and computation) are always more important than the other one in different applications. 

Thus, we need a set of FL evaluation metrics that are commonly agreed to be compatible with different scenarios, \ie, a set of \textit{standard} evaluation metrics. 
In other words, FL studies could be compared using these standard metrics under different scenarios with no ambiguity.

One good example of a compatible metric is the energy and carbon footprint \cite{flower-FL} since environmental wellness is one of the most important tasks of our society. FL models with fewer carbon emissions are better when achieving the same effectiveness.

\subsection{Real-time and Continuous Evaluations}

The evaluation of FL systems should be a real-time and continuous process. Specifically, the evaluation system should have the following functionalities:

\begin{itemize}
    \item \textit{Utility \& Efficiency Evaluation}: Requiring an easy-to-use evaluation interface and a group of benchmarking results (\eg, FL leaderboard). The system should enable researchers to evaluate new modes quickly, \eg, by uploading a simple script, and the system will automatically evaluate the new model. The evaluation results could be presented using a leaderboard, from which the researchers could quickly specify the state-of-the-art FL model and make performance comparisons.
    
    \item \textit{Security \& Privacy Evaluation}: Requiring a real-time and continuous verification to detect the attacks. Most of the FL studies use semi-honest security definitions, however, the security under the semi-honest assumption is not good enough for real-world applications because the parties that participated in the distributed training cannot fully trust each other, \ie, they will not believe that the others are semi-honest. Thus, real-time verification is essential to monitor each party's behavior and detect malicious participants deviating from the protocol. Furthermore, as we mentioned in section \Cref{sec:security and privacy evaluation}, private data leakage or model tampering may happen before, during, and after the FL training. Thus, security and privacy verification should be a real-time and continuous process.
\end{itemize}

\subsection{Contribution Evaluation for Incentive Design}

While not discussed in detail in this article, the incentive is also significant for FL, as parties work together only when incentives are designed satisfactorily. A suitable incentive mechanism in FL should satisfy the participants' rationality, meaning that each party's reward should be greater than the cost of joining the federation. Meanwhile, the parties with more contributions should gain more rewards to achieve fairness. There are also many other targets of designing an incentive mechanism for FL, such as reducing the delay in distributing rewards \cite{yu2020sustainable}. The evaluation plays a vital role in the incentive mechanism, especially when evaluating the participants' contributions. Intuitively, one participant's contribution could be evaluated by comparing the model performance when trained with and without its datasets, \eg, Shapley values \cite{wei2020efficient} is often adopted. The evaluation system could incorporate real-time contribution evaluation and reward distribution to serve as an incentive mechanism.

\subsection{Evaluation on FL platforms}

FL platforms are those frameworks that support simulating FL algorithms locally for research purposes or running FL in a distributed manner for industry applications. With the development of FL, many platforms have appeared: \eg, FATE \cite{FATE}, FedML \cite{he2020fedml}, FedScale \cite{fedscale}, \etc. However, in real-world applications or research studies, it is usually hard for users to determine which platform is the best choice under a certain scenario. Thus, evaluating these platforms is essential to benchmark and compare their efficiency and effectiveness under different scenarios. Meanwhile, we can also perform attack experiments on those platforms to assess privacy protection and uncover potential privacy issues before utilizing them in real-world applications. \hlminor{Notably, we can extend the evaluation goals and measures in this survey from evaluating algorithms into platforms, containing utility, security \& privacy, and efficiency. We discuss the extensibility of FedEval to evaluate different FL platforms in the appendix.}

%% file: sections/conclusion.tex
\section{Conclusion} \label{sec:conclusion}


\highlight{

In this survey, we provide a comprehensive overview of the evaluation goals and measures for FL studies. We categorized the key evaluation goals into utility, efficiency, and security \& privacy. For each goal, we reviewed commonly used metrics and evaluation methods from existing literature. We also discussed the necessity of conducting comprehensive evaluations across all goals due to the trade-offs between them. To facilitate such comprehensive analysis, we introduced FedEval, an open-source platform that simplifies implementing standardized FL evaluations. 

We also summarized several open challenges and future directions for FL evaluations. First, establishing standardized evaluation metrics that are compatible with different scenarios would enable fairer comparisons between different FL solutions. Second, developing capabilities for real-time verification of efficiency, utility, and especially security would be highly valuable for practical deployments. Third, evaluating the contributions of participants could support the design of incentive mechanisms. 

Overall, as FL continues maturing from the research domain towards real-world applications, strong evaluation methodologies will play an indispensable role in ensuring system quality and user trust. We hope this survey provides a useful reference for future efforts in advancing FL evaluation.

}